\documentclass{article}
\usepackage[final]{colm2026_conference}

\usepackage{url}

\usepackage{amsthm,amsmath}

\usepackage[english]{babel}
\usepackage[parfill]{parskip}
\usepackage{afterpage}
\usepackage{framed}
\usepackage{multirow}
\usepackage{arydshln}

\usepackage{graphicx}
\usepackage[labelfont=bf]{caption}
\usepackage{booktabs}

\usepackage{tcolorbox}
\tcbuselibrary{breakable}

\usepackage{listings}

\usepackage{listings}
\usepackage{fancyvrb}
\fvset{fontsize=\normalsize}

\usepackage{hyperref}
\hypersetup{colorlinks,
            linkcolor = blue,
            urlcolor  = blue,
            citecolor = blue,
            linktoc=all}

\usepackage[nameinlink]{cleveref}

\usepackage[acronym,nowarn,nohypertypes={acronym,notation}]{glossaries}
\setacronymstyle{long-sc-short}

\lstdefinestyle{mystyle}{
    commentstyle=\color{OliveGreen},
    keywordstyle=\color{BurntOrange},
    numberstyle=\tiny\color{black!60},
    stringstyle=\color{MidnightBlue},
    basicstyle=\ttfamily,
    breakatwhitespace=false,
    breaklines=true,
    captionpos=b,
    keepspaces=true,
    numbers=left,
    numbersep=5pt,
    showspaces=false,
    showstringspaces=false,
    showtabs=false,
    tabsize=2
}
\usepackage{tikz}
\usetikzlibrary{shapes,decorations,arrows,calc,arrows.meta,fit,positioning}
\tikzset{
    -Latex,auto,node distance =1 cm and 1 cm,semithick,
    state/.style ={circle, draw, minimum width = 0.7 cm},
    detstate/.style ={rectangle, draw, minimum width = 0.7 cm, minimum height = 0.7 cm},
    point/.style = {circle, draw, inner sep=0.04cm,fill,node contents={}},
    bidirected/.style={Latex-Latex,dashed},
    el/.style = {inner sep=2pt, align=left, sloped}
}
\usepackage{expl3}
\usepackage{siunitx}

\ExplSyntaxOn
\NewDocumentCommand{\speedup}{m o}{
  \IfValueTF{#2}
    {\textbf{\num{ \fp_eval:n { 256/(#1) } }}\,\ensuremath{\times}}
    {\num{ \fp_eval:n { 256/(#1) } }\,\ensuremath{\times}}
}
\ExplSyntaxOff

\usepackage{wrapfig}
\usepackage{mathtools}

\usepackage{mfirstuc}
\usepackage{xspace}

\makeatletter
\def\adl@drawiv#1#2#3{%
        \hskip.5\tabcolsep
        \xleaders#3{#2.5\@tempdimb #1{1}#2.5\@tempdimb}%
                #2\z@ plus1fil minus1fil\relax
        \hskip.5\tabcolsep}
\newcommand{\cdashlinelr}[1]{%
  \noalign{\vskip\aboverulesep
           \global\let\@dashdrawstore\adl@draw
           \global\let\adl@draw\adl@drawiv}
  \cdashline{#1}
  \noalign{\global\let\adl@draw\@dashdrawstore
           \vskip\belowrulesep}}
\makeatother

\usepackage{algorithm}
\usepackage{algorithmic}

\newacronym{smc}{\textsc{smc}}{sequential Monte Carlo}
\newacronym{mc}{\textsc{mc}}{masked conditional}

\newacronym{med}{\textsc{med}}{multi-token entropy decoding}

\newacronym{ntp}{\textsc{ntp}}{next-token prediction}
\newacronym{ntpar}{\textsc{ntp}}{next-token prediction}
\newacronym{ar}{\textsc{ar}}{auto-regressive}
\newacronym{dmc}{\textsc{dmc}}{diffusion Monte Carlo}
\newacronym{rlhf}{\textsc{rlhf}}{reinforcement learning from human feedback}
\newacronym{mdlm}{\textsc{mdlm}}{masked diffusion language model}
\newacronym{ips}{\textsc{ips}}{interacting particle system}
\newacronym{svdd}{\textsc{svdd}}{soft value-based decoding in diffusion models}

\newacronym{fk-ips}{\textsc{fk-ips}}{Feynman-Kac interacting particle system}
\newacronym{is}{\textsc{is}}{importance sampling}
\newacronym{tds}{\textsc{tds}}{twisted diffusion sampler}

\newacronym{fk}{\textsc{fk}}{Feynman-Kac}
\newacronym{fid}{\textsc{fid}}{Frechet inception distance}
\newacronym{ssim}{\textsc{ssim}}{structural similarity metric}

\newacronym{sde}{\textsc{sde}}{stochastic differential equation}
\newacronym{elbo}{\textsc{elbo}}{evidence lower bound}
\newacronym{kl}{\textsc{kl}}{Kullback-Leibler}

\newacronym{bpd}{\textsc{bpd}}{bits-per-dim}
\newacronym{vae}{vae}{variational autoencoder}

\newacronym{ode}{\textsc{ode}}{ordinary differential equation}
\newacronym{dbgm}{\textsc{dbgm}}{diffusion-based generative model}

\newacronym{vpsde}{\textsc{vpsde}}{variance-preserving stochastic differential equation}
\newacronym{vesde}{vesde}{variance-exploding stochastic differential equation}
\newacronym{alda}{alda}{accelerated Langevin diffusion}
\newacronym{malda}{malda}{modified accelerated Langevin diffusion}

\newacronym{mle}{mle}{maximum likelihood estimation}

\newacronym{cdf}{cdf}{cumulative density function}

\newacronym{hsm}{hsm}{Hybrid Score Matching}
\newacronym{dsm}{dsm}{Denoising Score Matching}
\newacronym{ism}{ism}{Implicit Score Matching}

\newacronym{iwae}{iwae}{importance-weighted auto-encoder}

\newacronym{vp}{vp}{variance preserving}
\newacronym{ve}{ve}{variance exploding}

\newacronym{mdm}{mdm}{Multivariate Diffusion Model}

\newacronym{pfode}{pfode}{\textit{probability flow} \gls{ode}}

\newacronym{fpe}{fpe}{Fokker-Planck equation}

\newtheorem*{assumption*}{Assumption}

\newtheorem*{corollary*}{Corollary}

\newtheorem*{definition*}{Definition}

\newtheorem*{lemma*}{Lemma}

\newtheorem*{proposition*}{Proposition}

\newtheorem*{theorem*}{Theorem}

\newcommand{\qdata}{q_{\text{data}}}

\newcommand{\kl}[1]{\textsc{KL}\left(#1\right)}

\renewcommand{\mid}{~\vert~}

\newcommand{\mba}{\mathbf{a}}

\newcommand{\mbc}{\mathbf{c}}

\newcommand{\mbo}{\mathbf{o}}

\newcommand{\mbr}{\mathbf{r}}

\newcommand{\mbx}{\mathbf{x}}

\newcommand{\mbE}{\mathbf{E}}

\newcommand{\cL}{\mathcal{L}}

\usepackage{nicefrac}
\usepackage{subcaption}
\usepackage{wrapfig}
\usepackage{titlesec}

\usepackage[utf8]{inputenc}
\usepackage[T1]{fontenc}
\usepackage{url}
\usepackage{booktabs}
\usepackage{nicefrac}
\usepackage{xcolor}
\usepackage{soul}
\usepackage{float}

\usepackage{url}
\usepackage{enumitem}
\usepackage{natbib}

\usepackage{lineno}

\title{Rethinking Reasoning with Masked Diffusion Language Models: Early Exits, Post-hoc Reasoning, and Beyond}

\author{Zachary Horvitz$^{*,1}$, Raghav Singhal$^{*,2}$, Hao Zou$^{1}$, Carles Domingo-Enrich$^{3}$ \\
\textbf{Zhou Yu$^{1}$, Rajesh Ranganath$^{2, 4}$, Kathleen McKeown$^{1}$ } \\
\textsuperscript{1}Columbia University \qquad \textsuperscript{3}Microsoft Research \\ 
\textsuperscript{2}Department of Computer Science, New York University \\ 
\textsuperscript{4}Center for Data Science, New York University
}

\begin{document}

\ifcolmsubmission
\linenumbers
\fi

\maketitle

\begin{abstract}

The reasoning paradigm, where language models reason before answering, has enabled breakthroughs on tasks such as mathematical problem-solving. While current tooling for reasoning is built around next-token prediction trained models, recent works introduce an alternative choice: \glspl{mdlm}. \Glspl{mdlm} are trained to in-fill positions in randomly masked sequences. We introduce \textit{reasoning-as-infilling}, a prompting technique that pre-fills tokens to explicitly delimit reasoning and answer regions, unlocking a unified set of capabilities for \gls{mdlm} reasoning. Because answer positions are explicitly designated, the model's conditional distributions over answer tokens are directly accessible during generation. This enables \textit{early exits} when the model is certain of its answer. The same framework supports \textit{post-hoc reasoning}: given question-answer pairs, \glspl{mdlm} can sample high-quality reasoning traces from their posterior, a distribution that is intractable for autoregressive models. On GSM8k, fine-tuning LLaDA-8B-Base on these posterior traces improves accuracy by $+14.9\%$, matching gains from human-written traces ($+13.4\%$). Finally, given a reference answer, the answer region distributions enable \textit{scoring partial reasoning traces} at \textit{intermediate steps}, providing intermediate rewards that are more strongly correlated with correctness than scores from a specialized process reward model. At intermediate steps, answer-likelihood scores from autoregressive models are significantly less predictive of correctness than those from MDLMs. Our results demonstrate that the \gls{mdlm} training objective provides promising benefits for reasoning.

\end{abstract}

\section{Introduction}
\def\thefootnote{*}\footnotetext{Denotes equal authorship. Correspondence to \url{zfh2000@columbia.edu}, \url{rsinghal@nyu.edu}.}

The current dominant approach for language modeling is based on \gls{ntpar} training. \Gls{ntp} language models learn the conditional distribution of the \textit{next token} given the previous tokens in a sequence \citep{shannon1951prediction,radford2019language}. The resulting language model is sampled auto-regressively left-to-right, one token at a time. Recent work proposes \glspl{mdlm} \citep{austin2021structured,sahoo2024simple,shi2024simplified} as an alternative to \gls{ntpar} models.  \glspl{mdlm} are trained to in-fill sequences with randomly masked positions. The resulting model learns the distributions $p_\theta(x^i | \mbx_{\textsc{un-masked}})$ at every masked position $i$. 

\begin{figure*}[ht]
    \centering
    \includegraphics[width=0.45\linewidth]{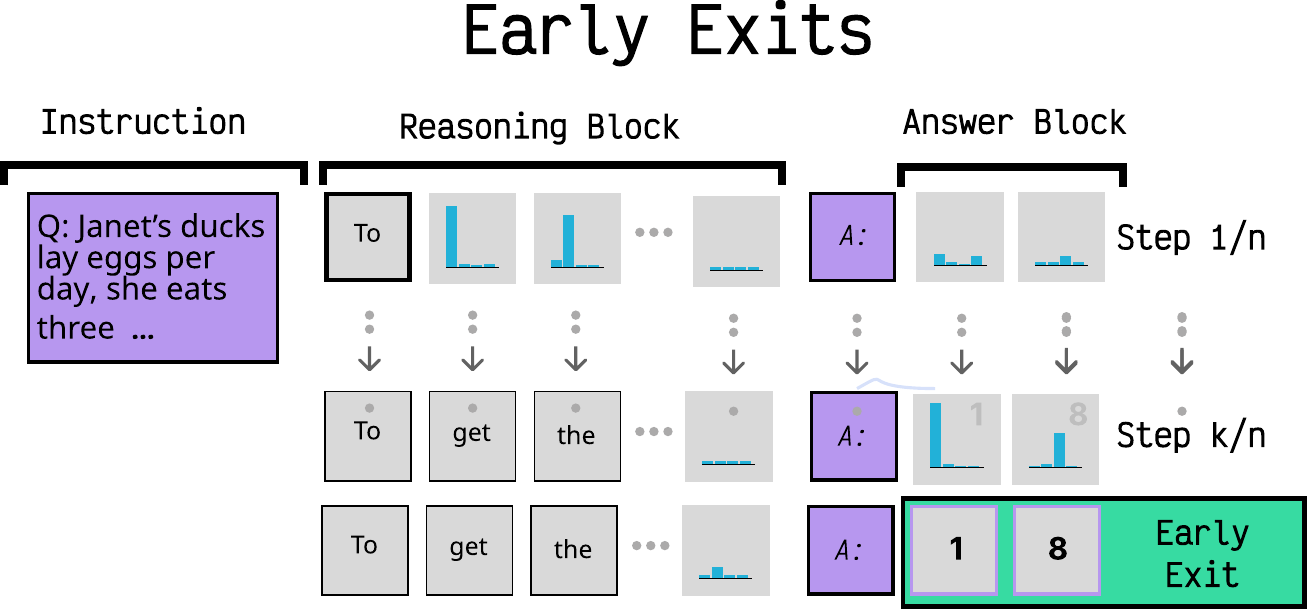}
  \hspace{0.05\linewidth}
  \includegraphics[width=0.45\linewidth]{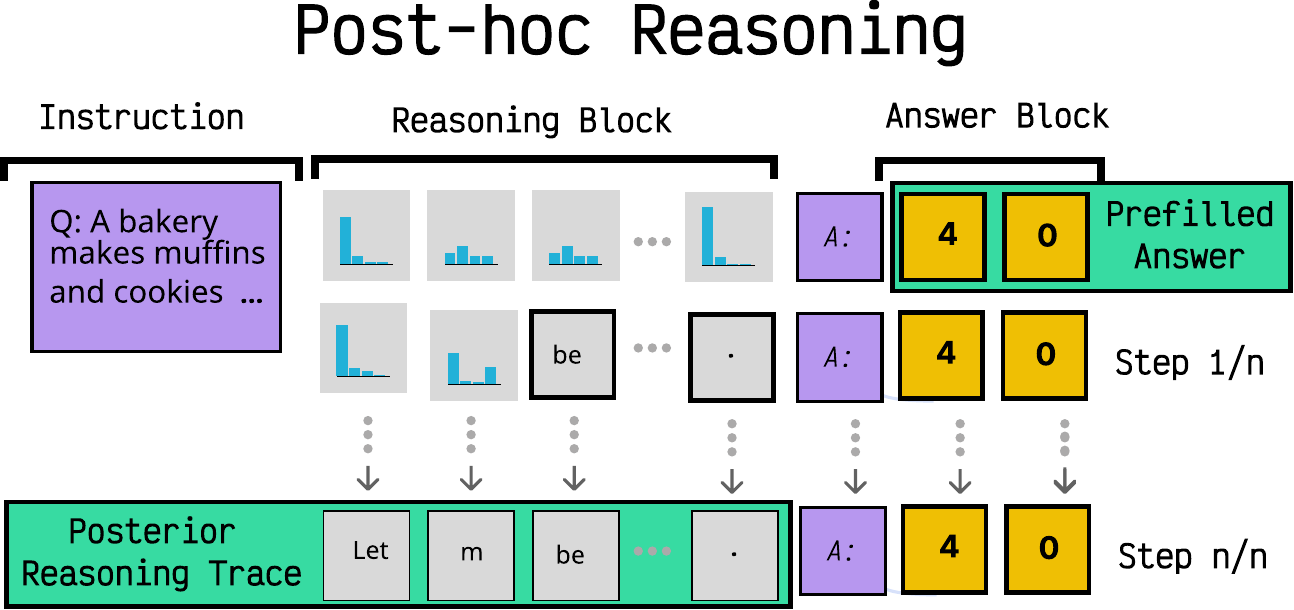}
    \caption{\glspl{mdlm} learn the conditional distributions at each masked token position. We reframe reasoning as infilling a prompted reasoning template, which enables directly modeling answer token probabilities \textit{during} reasoning. This provides several benefits, like enabling early exits (\textit{left panel}) or \textit{post-hoc} reasoning given a pre-filled answer (\textit{right panel}). 
    }
    \label{fig:reasoning_template}
\end{figure*}
\begin{figure*}[t]
    \centering
    \includegraphics[width=1\linewidth]{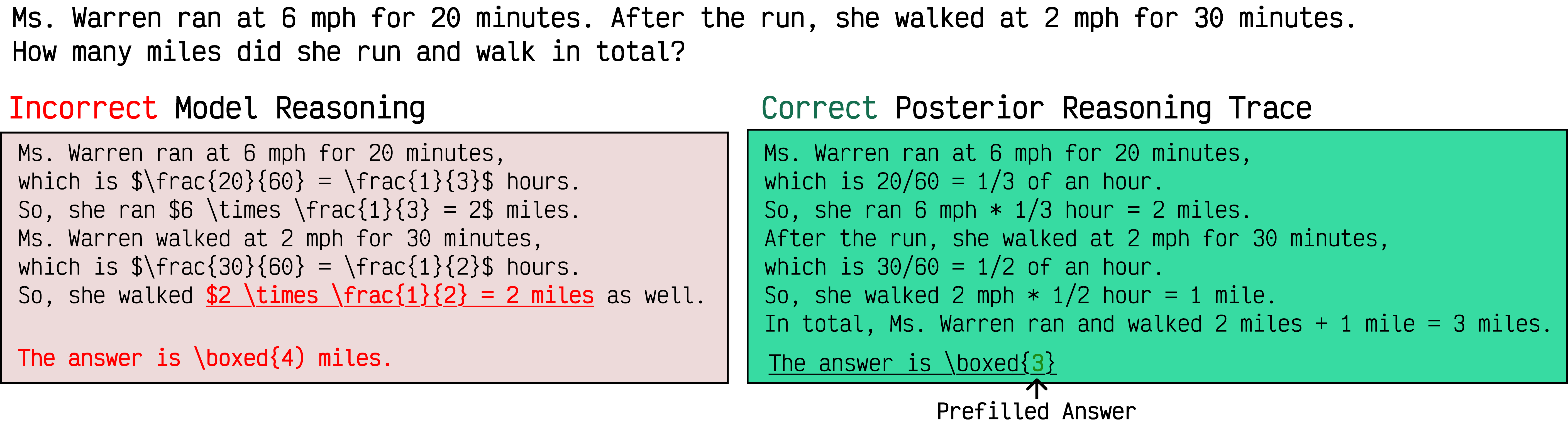}
    \caption{\textbf{Posterior reasoning from correct answers is a source of high-quality reasoning traces.} Pre-filling a correct answer into a reasoning template elicits correct reasoning for a large portion of problems that the model initially fails to solve. For example, here greedily sampling from LLaDA-8B-Instruct \citep{nie2025largelanguagediffusionmodels} yields an incorrect reasoning step that is corrected in the posterior reasoning trace.}
    \label{fig:reasoning_as_infilling}
\end{figure*}

While modeling all masked positions requires additional effort, \glspl{mdlm} have several potential benefits, such as parallel token decoding \citep{sahoo2024simple,sahoo2025esoteric, ben2025accelerated, wu2025fast}, and flexible decoding orders \citep{kim2025train} that lead to significant improvements on logic puzzles, such as Sudoku. Additionally, \citet{bachmann2024pitfalls,prabhudesai2025diffusion}
show that multi-token prediction objectives can achieve better likelihoods and accuracy on tasks,
and access to the distribution and samples from masked positions supports controllable generation \citep{schiff2024simple,singhal2025generalframeworkinferencetimescaling}.

In this work, we demonstrate that the conditional distributions of all masked positions, together with the ability of \glspl{mdlm} to in-fill, unlock new sampling and post-training capabilities for reasoning that are not readily available for \gls{ntp} models.

We propose \textbf{{reasoning-as-infilling}}, where we add user-specified contexts in multiple positions, not just the beginning of the sequence, unlike \gls{ntp} models. Specifically, we pre-fill an explicit reasoning template that delimits specific reasoning and answer positions (see \cref{fig:reasoning_template}). This single change yields multiple inference-time and post-training capabilities.

First, because the answer positions are explicitly designated, the \gls{mdlm}'s conditional distributions over answer tokens are directly accessible \textit{during} generation. This enables \textbf{early exits}: once the model's answer-region entropy falls below a threshold, the remaining reasoning steps can be skipped. On \textsc{gsm8}k, early exits alone yield a $1.3\times$ acceleration with less than a $1$ percentage point reduction in accuracy, and can be combined with parallel decoding algorithms for additional acceleration. 
Reasoning-as-infilling also integrates naturally with \textit{variable-sequence length} diffusion approaches \citep{Dreamon2025,havasi2025edit,kim2025anyorderflexiblelengthmasked}, enabling flexible reasoning budgets at inference time.

Second, the same framework directly supports post-training. Given a correct answer, pre-filling the answer block enables sampling from the \gls{mdlm}'s posterior $p_\theta(\mbr \mid \mbc, \mba)$, a distribution that is intractable for \gls{ntp} models but trivial for \glspl{mdlm}. This yields \textbf{high-quality post-hoc reasoning traces} that significantly improve model performance when used for fine-tuning, matching the gains from expensive human-annotated traces.

Third, given a reference answer, the same answer-conditioned distributions provide a built-in mechanism for \textbf{scoring partial reasoning traces} at intermediate steps, without training an auxiliary process reward model. The marginal probabilities of the answer block correlate more strongly with final output correctness than scores from a specialized 7 billion parameter process reward model \citep{zhang2025lessons}, suggesting applications for steering \citep{schiff2024simple,singhal2025generalframeworkinferencetimescaling} and filtering during post-training.

\paragraph{Contributions.} We introduce {reasoning-as-infilling}, a method that leverages \gls{mdlm}'s infilling capabilities to explicitly distinguish reasoning and answer tokens. This framework naturally provides three key benefits:
\begin{itemize}[leftmargin=*]
    \item \textit{Early exits}, where we skip the remaining reasoning steps once the model is sufficiently certain of the answer. On \textsc{gsm8}k with one-token decoding, this leads to a \textit{$1.3\times$ acceleration with less than a $1$ percentage point reduction in accuracy}. Early exits can be combined with parallel decoding algorithms for additional acceleration.
    
    \item \textit{Post-hoc reasoning}, where given question-answer pairs, we generate reasoning traces conditioned on the answer. On the \textsc{gsm8}k dataset, supervised fine-tuning on these reasoning traces can improve the model (+14.9\%) as much as supervised fine-tuning on the expensive \textit{human-annotated} \textsc{gsm8}k reasoning traces (+13.4\%).    Posterior traces are also judged more correct than traces from STaR-style prefix-hint prompting for the same models.  
            
    \item \textit{Scoring reasoning traces}, where given a question-answer pair, we can score the reasoning process for correctness at intermediate steps using the distributions of the answer block. These scores correlate with whether the reasoning process will produce correct answers.  In contrast, we find that analogous answer likelihoods computed by the evaluated auto-regressive models are only weakly informative until late in generation.

\end{itemize}

\section{Related Work}

\paragraph{Early Exits.} Concurrently, \citet{li2025diffusion} demonstrate that \glspl{mdlm} assign high probability to correct answers before decoding a full reasoning trace. They also propose a method for decoding answers early, based on the model's top logits over a pre-specified region of the sequence. In one configuration of their method, similar to \textit{reasoning-as-infilling}, they pre-fill a suffix to delimit an answer region. Our work differs in several respects: we use an upper-bound on answer region entropy as the criterion for certainty, rather than marginal probability scores, and we ensure the answer region is sampled from the model's joint distribution rather than decoded in one step. Beyond early exits, we demonstrate that the reasoning-as-infilling framework extends naturally to parallel decoding and variable-length decoding algorithms \citep{kim2025anyorderflexiblelengthmasked, Dreamon2025, havasi2025edit}, further accelerating inference and enabling flexible reasoning budgets. Critically, we go beyond early exiting to show that the same infilling framework supports post-hoc reasoning trace generation and intermediate scoring.

\paragraph{Post-hoc Reasoning and Scoring.}
\citet{zelikman2022star} proposes fine-tuning language models on reasoning traces that are generated conditioned on a correct answer. \citet{phan2023training,ruan2025reasoning} propose fine-tuning a model on samples from approximations of the posterior $p_\theta(\mbr \mid \mbc, \mba)$. In contrast, \glspl{mdlm} enable sampling from the posterior of the reasoning traces given the answer by simply in-filling the answer in the answer block provided in the {reasoning-as-infilling} framework. For \glspl{mdlm}, \citet{zhao2025inpainting} propose in-filling guided policy optimization, where they randomly mask and infill human-written reasoning traces during training on problems where the model does not generate a correct reasoning trace and answer. In contrast, we sample from \glspl{mdlm} given only the gold answer and no human-written reasoning traces. 

Using log-probabilities for scoring or training has been explored for autoregressive models \citep{liu2025nover,zhou2505reinforcing}. In contrast to auto-regressive models, \glspl{mdlm} provide access to the conditional distributions of the answer tokens at \textit{every intermediate step} of the reasoning process, as a byproduct of each forward pass. This yields dense, step-level scores, without requiring completed rollouts or training a separate process reward model \citep{zhang2024rest,silver2016mastering}.

\section{Masked Diffusion Language Models}
\Glspl{mdlm} \citep{sohl2015deep,devlin2019bert,austin2021structured,sahoo2024simple,shi2024simplified} are a class of generative models for modeling discrete data $\mbx \sim \qdata$, where $\mbx = (x^1, x^2, \dots, x^{L})$ and each position $x^i$ takes values in a finite vocabulary $\mathcal{V}$. For training the model $p_\theta(\mbx)$, the sequence $\mbx \sim \qdata$ is masked randomly and the model learns to predict the distributions of the masked positions for a fixed length sequence by maximizing a lower bound on the log-likelihood of $\mbx$:
\begin{align}\nonumber
 \cL(\mbx, \theta): = \mathbf{E}_{\textsc{mask} \sim U}  \sum_{j \in \textsc{mask}}  \log p_\theta( x^j \mid \mbx_{\textsc{un-masked}})
\end{align}
where $\textsc{masked-set}$ is a set of randomly masked positions in $\{1, 2, \dots, L\} $. Sampling from an \gls{mdlm} is performed by iteratively un-masking positions. Notably, an \gls{mdlm} can also  be viewed as any-order auto-regressive model \citep{uria2014deep,ou2024your}, where given a decoding order $\mbo = [o_1, \dots, o_L]$ with $o_j \in \{1, 2, \dots, L \}$, the model likelihood can be defined as:
\begin{align}\label{eq:anyorder_model}
    p_\theta(\mbx \mid \mbo) = \prod_{j=0}^{L-1} p_\theta(x^{o(j)} \mid \mbx_{o(<j)}) 
\end{align}
where $o_j$ refers to the position decoded at step $j$ and $\mbx_{o(<j)}$ refers to all positions decoded prior to step $j$. 

Sampling from an \gls{mdlm} model requires defining a decoding order. Popular sampling  approaches for \glspl{mdlm} select positions to unmask based on confidence (e.g. token probability \citep{chang2022maskgit,wu2025fast} or entropy \citep{kim2025train,ye2025dream,ben2025accelerated}). These sampling algorithms are often combined with block sampling methods \citep{sahoo2024simple,arriola2025blockdiffusioninterpolatingautoregressive},  which define a left-to-right sequence of fixed length blocks and decode within each block in  an arbitrary order \citep{sahoo2024simple,nie2025largelanguagediffusionmodels,arriola2025blockdiffusioninterpolatingautoregressive}.

Recently, several works \citep{Dreamon2025,havasi2025edit,kim2025anyorderflexiblelengthmasked} have proposed variable-sequence \glspl{mdlm}. Instead of unmasking a fixed-length sequence, these methods dynamically resize the output at inference-time using expand and delete operations.

\paragraph{What do \glspl{mdlm} offer for reasoning?}
One benefit of \glspl{mdlm} for reasoning tasks is their ability to decode tokens in parallel \citep{ben2025accelerated,wu2025fast}, accelerating inference and data generation. Another possible benefit is their ability to decode sequences in arbitrary orders. For instance, \citet{kim2025train} show that any-order decoding outperforms next-token prediction on puzzles such as Sudoku.
However, our experiments show that any-order decoding offers \textit{limited benefits} for reasoning tasks. Naive or block any-order decoding algorithms either sample a large portion of tokens in a left-to-right order (nearly $50\%$ for \textsc{gsm8k}) or underperform left-to-right sampling (see \cref{fig:decoding_strategies}). Qualitatively, any-order decoding often elicits a surprising behavior: \glspl{mdlm} decode an incorrect answer first, followed by a post-hoc rationale (see \cref{appsec:observations} for additional analysis).

This ``pathological'' behavior is revealing: it demonstrates that \glspl{mdlm} naturally support many different factorizations of the joint distribution $p_\theta(\mba, \mbr \mid \mbc)$ over reasoning traces $\mbr$, input context $\mbc$, and answers $\mba$. At inference-time, sampling a possibly incorrect answer $\mathbf{\hat{a}}$ before a reasoning trace is undesirable. However, when the gold answer is known, as in post-training, this same factorization becomes a valuable source of reasoning data. Moreover, the tendency of \glspl{mdlm} to converge on answers early suggests that models may be certain of an answer well before completing their full reasoning trace. These observations motivate \textit{reasoning-as-infilling}.

\section{Rethinking Reasoning with MDLMs}

\glspl{mdlm} are prompted similarly to \gls{ntp} models, and the distributions of the masked positions are used only for sampling high-confidence tokens. The remaining distributions are discarded. We show that the ability of \glspl{mdlm} to in-fill and to access the distribution of \textit{all} masked positions unlocks new inference-time and post-training capabilities.

\subsection{Reasoning-as-Infilling with MDLMs}
Generally, \gls{ntp} models are controlled at inference-time with a prompt prefix $\mbc$ that is inserted at the beginning of the sequence. However, for \glspl{mdlm} we propose pre-filling the output sequence with user-specified tokens. In the case of reasoning tasks, where a model produces a reasoning trace prior to answering, we can pre-fill the output sequence with a reasoning template that distinguishes the reasoning and answer token positions:

 \begin{align}\label{eq:reasoning_as_infilling}\nonumber
\left[\begin{array}{l}
\underbrace{[\textsc{m}]_1  \text{ } [\textsc{m}]_2  \text{ }  \dots  \text{ }  [\textsc{m}]_k}_{\text{reasoning block}}  \text{ }  \text{<Delimiter>}  \text{ }  
\underbrace{[\textsc{m}]_{k+1}  \text{ }  \dots  \text{ }  [\textsc{m}]_L}_{\text{answer block}}
\end{array}\right]
\end{align}
Here the answer delimiter is a user-specified choice (e.g. \textit{"The answer is: "} for math tasks, or function definitions for a coding task). In this reformulation of prompting, the context $\mbc$ now includes both the prompt and the answer delimiter, see 
\cref{fig:reasoning_template}. 
By distinguishing between reasoning and answer positions, reasoning-as-infilling offers several advantages for sampling and post-training.

\paragraph{Early exits.}
By designating explicit answer block positions, reasoning-as-infilling enables measuring answer uncertainty \textit{while generating the reasoning trace}. A measure of uncertainty is the entropy of the answer block given the unmasked reasoning positions. This joint entropy requires additional estimation as \glspl{mdlm} only provide access to the marginals
$p_\theta( a_j \mid \mbr_\textsc{unmasked}, \mbc)$. However, we show that the marginal distributions can be used to upper-bound the joint entropy,
\begin{align}
   H_{\textsc{ub}} &:= \sum_{j \in \textsc{answer-block}} H(a_j \mid \mbr_{\textsc{unmasked}}, \mbc) \geq H(\mba \mid \mbr_{\textsc{unmasked}}, \mbc)
\end{align}
See \cref{appsec:proofs} for a proof. Using this quantity, we propose \textbf{early exiting based on the answer uncertainty upper-bound} $H_\textsc{ub}$. That is, given a partial reasoning trace, $\mbr_\textsc{unmasked}$, we skip filling in the remaining reasoning tokens if the answer-entropy upper bound falls below a user-specified threshold $\gamma$, $H_\textsc{ub} < \gamma$.

\paragraph{Post-training \glspl{mdlm} with reasoning-as-infilling.}

Typically, post-training a model to reason uses expensive human demonstrations \citep{ouyang2022traininglanguagemodelsfollow}. Alternatively, \citet{zelikman2022star,phan2023training,ruan2025reasoning} have demonstrated that post-training on model generated reasoning traces provides an alternative for improving performance \citep{zelikman2022star,phan2023training,ruan2025reasoning}. 

These methods work off the principle that sampling reasoning traces from the posterior $p_\theta(\mbr \mid \mbc, \mba)$ and then training on these samples
can increase the likelihood of generating correct answers. However, sampling from the posterior $p_\theta(\mbr \mid \mba, \mbc)$ is intractable for \gls{ntp} models: the autoregressive factorization generates tokens left-to-right, making it impossible to condition on answer tokens that appear later in the sequence without either significant prompt engineering or training another model to yield reasoning traces given answer hints \citep{phan2023training,zelikman2022star}.

Because \glspl{mdlm} model the conditional distributions of \textit{all} masked positions given any set of unmasked tokens, conditioning on the answer is as simple as pre-filling the answer block positions. With {reasoning-as-infilling}, one can \textit{simply} pre-fill the answer positions to sample directly from the posterior distribution, without prompt engineering or auxiliary models. These posterior traces can be used for post-training in several ways, including with STaR \citep{zelikman2022star}, maximum marginal log-likelihood training \citep{phan2023training,pml2Book}, or maximizing the likelihood on the answer and the posterior reasoning traces: $\max_\theta \sum_{i=1}^N \log p_\theta(\mba_i, \mbr_i \mid \mbc_i)$ where the posterior reasoning traces are generated by the model, $\mbr_i \sim p_\theta(\mbr_i \mid \mbc_i, \mba_i)$.

\paragraph{Scoring partial reasoning traces when post-training.} 
Existing fine-tuning algorithms, such as \textsc{grpo} \citep{shao2024deepseekmath} and \textsc{rloo} \citep{ahmadian2024back}, do not make use of posterior samples but score the generations upon completion. These algorithms can benefit from intermediate rewards \citep{silver2016mastering}. Recent work has shown that guiding the generation process with intermediate rewards produces samples that improve model fine-tuning \citep{zhang2024rest}. These intermediate rewards are generally provided by an \textit{external} pre-trained process reward model (PRM) \citep{lightman2023letsverifystepstep,zhang2024rest, zhang2025lessonsdevelopingprocessreward}. 

Given the answer, reasoning-as-infilling enables \glspl{mdlm} to score arbitrary reasoning traces at intermediate steps. This is possible because \glspl{mdlm} jointly model all positions, including the answer block distributions, as part of each forward pass during generation. Given a partial reasoning trace $\mbr_{\textsc{unmasked}}$ and an answer $\mba^*$, we score $\mbr_{\textsc{unmasked}}$ with:
\begin{align}\label{eq:iprm}
    & \phi_{\textsc{iprm}}(\mbr_{\textsc{unmasked}} \mid \mbc, \mba^*)   := \sum_{j \in \textsc{answer-block}} \log p_\theta(a_j = a_j^* \mid \mbc, \mbr_{\textsc{unmasked}}) .
\end{align}
The intuition behind the equation is that when the likelihood of individual answer tokens is higher for the reasoning trace $\mbr_{\textsc{unmasked}}$, then $\mbr_{\textsc{unmasked}}$ is often more likely to produce the answer.  Unlike the entropy criterion used for early exits, which requires no answer, $\phi_\textsc{iprm}$ assumes a reference or verifier-provided answer $\mba^*$, and is primarily applicable to post-training settings where labels are available.

\begin{table}[t]
    \centering
    \small
    \resizebox{\linewidth}{!}{%
    \begin{tabular}{ll ccc ccc}
\toprule
\multicolumn{2}{c}{} & \multicolumn{3}{c}{\textbf{GSM8K}} & \multicolumn{3}{c}{\textbf{Math500}} \\
\cmidrule(lr){3-5} \cmidrule(lr){6-8}
\textbf{Model} & \textbf{Sampler} & \textbf{Exit Param} & \textbf{Speed-up} $\uparrow$ & \textbf{Acc.} & \textbf{Exit Param} & \textbf{Speed-up} $\uparrow$ & \textbf{Acc.}  \\
\midrule
LLaDA & {\textsc{entropy}, $k=1$} & \textsc{no template} & \speedup{256} & 76.6 & \textsc{no template} & \speedup{256} & 33.8 \\
LLaDA & {\textsc{entropy}, $k=1$} & \textsc{no exit}     & \speedup{256} & 79.4 & \textsc{no exit}     & \speedup{256} & 33.4 \\
LLaDA & \textsc{entropy}, $k=1$   & $\gamma=0.1$         & \speedup{193} & 78.6 & $\gamma=0.3$         & \speedup{221} & 31.9 \\
\midrule
LLaDA & \textsc{med}, $\lambda = 0.2$ & \textsc{no exit} & \speedup{94} & 79.9 & \textsc{no exit} & \speedup{143} & 33.4 \\
LLaDA & \textsc{med}, $\lambda = 0.2$ & $\gamma=0.1$     & \speedup{77} & 79.3 & $\gamma=0.3$     & \speedup{129} & 32.0 \\
\midrule
LLaDA & \textsc{eb}-sampler, $\lambda = 0.1$ & \textsc{no exit} & \speedup{93} & 79.5 & \textsc{no exit} & \speedup{118} & 34.6 \\
LLaDA & \textsc{eb}-sampler, $\lambda = 0.1$ & $\gamma=0.1$     & \speedup{74} & 78.7 & $\gamma=0.3$     & \speedup{107} & 33.6 \\
\midrule
LLaDA & \textsc{bpcd}, $\lambda = 0.9$ & \textsc{no exit} & \speedup{78} & 79.0 & \textsc{no exit} & \speedup{95} & 33.2 \\
LLaDA & \textsc{bpcd}, $\lambda = 0.9$ & $\gamma=0.1$     & \textbf{[\speedup{64}[bf]]} & 78.9 & $\gamma=0.3$     & \textbf{[\speedup{86}]} & 31.8 \\
\midrule\midrule
Dream & {\textsc{entropy}, $k=1$} & \textsc{no template} & \speedup{256} & 80.1 & \textsc{no template} & \speedup{256} & 33.4 \\
Dream & {\textsc{entropy}, $k=1$} & \textsc{no exit}     & \speedup{256} & 79.8 & \textsc{no exit}     & \speedup{256} & 35.6 \\
Dream & {\textsc{entropy}, $k=1$} & $\gamma=0.7$         & \speedup{225} & 76.7 & $\gamma=0.7$         & \speedup{245} & 33.2 \\
\midrule
Dream & \textsc{med}, $\lambda = 0.2$ & \textsc{no exit} & \speedup{135} & 79.2 & \textsc{no exit} & \speedup{147} & 35.6 \\
Dream & \textsc{med}, $\lambda = 0.2$ & $\gamma=0.7$     & \speedup{121} & 79.3 & $\gamma=0.7$     & \speedup{141} & 35.4 \\
\midrule
Dream & \textsc{eb}-sampler, $\lambda = 0.1$ & \textsc{no exit} & \speedup{128} & 77.6 & \textsc{no exit} & \speedup{135} & 32.2 \\
Dream & \textsc{eb}-sampler, $\lambda = 0.1$ & $\gamma=0.7$     & \speedup{115} & 75.3 & $\gamma=0.7$     & \textbf{[\speedup{75}]} & 30.2 \\
\midrule
Dream & \textsc{bpcd}, $\lambda = 0.9$ & \textsc{no exit} & \speedup{115} & 78.0 & \textsc{no exit} & \speedup{116} & 32.0 \\
Dream & \textsc{bpcd}, $\lambda = 0.9$ & $\gamma=0.7$     & \textbf{[\speedup{103}]} & 74.6 & $\gamma=0.7$     & \speedup{104} & 30.8 \\
\bottomrule
    \end{tabular}%
    }
    \caption{\textbf{Early exits accelerate \gls{mdlm} inference.} We evaluate LLaDA-8B-Instruct \cite{nie2025largelanguagediffusionmodels} and Dream-7B-Instruct \citep{dream2025} on \textsc{gsm8}k. We report acceleration relative to full-length generation with a sequence length of $256$. Varying the early-exit threshold $\gamma$ trades faster inference for task accuracy. Early exits can be combined with parallel decoding algorithms for additional acceleration.}
    \label{tab:early_exit_combined}
\end{table}
\section{Experiments}
In this section, we examine the inference-time and post-training benefits of {reasoning-as-infilling}, such as \textbf{early-exits} based on answer certainty, the ability to \textbf{bootstrap reasoning traces} from correct answers, and their intrinsic ability to \textbf{score partial reasoning traces} given (question, answer) pairs.

\begin{table}[t]
\centering
\small
\resizebox{\linewidth}{!}{%
\begin{tabular}{llcccc}
\toprule
\multicolumn{2}{l}{} & \multicolumn{4}{c}{\textbf{GSM8K}} \\
\cmidrule(lr){3-6}
\textbf{Model} & \textbf{Setting} & \textbf{Exit Param} & \textbf{Speed-up} $\uparrow$ & \textbf{Seconds$\downarrow$} & \textbf{Acc.} \\
\midrule
\multirow{2}{*}{DreamOn}
& \textsc{No Template} (Fixed Length) & \textsc{No Exit} & \speedup{256} & 10.8 & \textcolor{red}{1.4\%} \\
& \textsc{No Template} & \textsc{No Exit} & \speedup{256} & 8.7 & \textcolor{red}{4.9\%} \\
\midrule
\multirow{4}{*}{DreamOn w/ Template}
& \textsc{With Template} (Fixed Length) & \textsc{No Exit} & \speedup{256} & 11.0 & 60.6\% \\
& \textsc{With Template} & \textsc{No Exit} & \speedup{256} & 10.3 & 66.4\% \\
\cmidrule(lr){2-6}
& \textsc{With Template} (Fixed Length) & $\gamma=0.5$ & \speedup{137} & 5.9 & 64.7\% \\
& \textsc{With Template} & $\gamma=0.5$ & \speedup{153} & 6.5 & 66.0\% \\
\bottomrule
\end{tabular}%
}
\caption{\textbf{Reasoning-as-infilling can be used alongside variable-length diffusion models.} We evaluate DreamOn \citep{Dreamon2025} on \textsc{gsm8}k. We consider both fixed-length decoding, and utilizing deletion operation to dynamically resize the reasoning budget. We report accuracy, along with speed-up relative to full-length generation ($256$ steps). We also report wall-clock seconds on an Nvidia A100 GPU. Notably, while DreamOn was fine-tuned for variable-length generation on a coding dataset \citep{Huang2024OpenCoderTO}, reasoning-as-infilling elicits correct mathematical reasoning.}
\label{tab:dreamon_gsm8k_earlyexit}
\end{table}

\subsection{Early exits}
We investigate the inference-time benefits of {reasoning-as-infilling} on two mathematical reasoning benchmarks, \textsc{gsm8}k \citep{cobbe2021training} and \textsc{math500} \citep{lightman2023letsverifystepstep}, with the Dream-7B-Instruct and LLaDA-8B-Instruct models. For both tasks, we consider a generation length $L = 256$ with block size 32. We pre-fill the answer delimiter \textit{``The answer is \textbackslash boxed\{.."}, and allocate $10$ answer tokens. As a baseline, we compare against allocating a sequence of length $256$ with no-reasoning template. 

\paragraph{Early Exiting Alongside Multi-token Decoding.} For sampling, we examine early exits with one-token decoding along with recent multi-token decoding methods. We use entropy-bound (\textsc{eb}) sampler \citep{ben2025accelerated}, block parallel confidence decoding (\textsc{bpcd}) \citep{wu2025fast}, as well as a simple entropy-based multi-token method that we propose called \textsc{med}, which we describe in \cref{appsec:med}. For a comparison of the three samplers, see \cref{fig:multi_token_kl} in \cref{appsec:med}.

In \cref{tab:early_exit_combined}, we observe:
\begin{itemize}[leftmargin=*]
    \item \textbf{Faster inference with minimal accuracy degradation:} For LLaDA, early exiting yields a $1.3\times$ acceleration on one-token entropy decoding with only a $0.8$ percentage point drop in performance versus the baseline reasoning template. 
    
    \item  \textbf{Benefits compound with multi-token decoding}. Early exits combined with multi-token decoding provide further acceleration. For LLaDA on \textsc{gsm8}k, \textsc{bpcd} alone yields a $3.3\times$ acceleration. Adding early exits yields a $4.0\times$ total acceleration, a further $1.2\times$ acceleration over \textsc{bpcd} alone, while accuracy changes from $79.0$ to $78.9$. In \cref{appsec:exits}, we provide examples of reasoning traces with varying exit thresholds. 
\end{itemize}

Notably, the benefits of early exits are more pronounced for LLaDA than Dream, which requires higher exit thresholds for speed-ups. This may be due to Dream's adaptation from an \gls{ntp} model \citep{gong2024scaling,dream2025}. See \cref{appsec:observations} for a discussion of the sampling behavior of Dream and LLaDA.

We also compare entropy-based exits with Prophet \citep{li2025diffusion} on GSM8k and CommonsenseQA. On GSM8k, the methods are matched at comparable NFE budgets (74.8\% vs. 74.3\% at ~1.55$\times$). On CommonsenseQA, entropy exits are both more accurate and faster at every evaluated threshold (for example, $76.2\%$ at $1.86\times$ vs. $72.2\%$ at $1.02\times$). Full sweeps are in \cref{appsec:prophet}.

\paragraph{Early Exiting with Variable-Length Decoding.} One potential drawback of reasoning-as-infilling with \glspl{mdlm} is that it pre-defines a fixed reasoning trace length. We next examine how reasoning-as-infilling and early exits can be utilized alongside variable-length diffusion models \citep{Dreamon2025,havasi2025edit,kim2025anyorderflexiblelengthmasked}. 
We evaluate DreamOn-7B, which was fine-tuned for variable-length decoding on a dataset of coding problems \citep{Huang2024OpenCoderTO}, on \textsc{gsm8}k with an initial sequence length of $256$ and block size $32$ decoding. We consider both fixed-length decoding and sampling with dynamic reasoning-trace lengths via deletions.

In \cref{tab:dreamon_gsm8k_earlyexit}, reasoning-as-infilling significantly improves the model's mathematical reasoning, eliciting correct reasoning despite the model's fine-tuning on coding tasks. Without reasoning-as-infilling DreamOn-7B performs poorly on \textsc{gsm8}k, with an accuracy of $1.4\%$. Enabling deletions further improves accuracy and provides a 6\% reduction in sampling time.\footnote{We report average sampling time on an Nvidia A100 40GB GPU. DreamOn deletions count towards NFEs, but reduce the length of the sequence, and memory requirements of each forward pass.} Incorporating early exits provides an additional $37\%$ reduction in sampling time while maintaining accuracy.

\begin{table}[t]
\centering
\begin{tabular}{lcc}
\toprule
\textbf{Model} & \multicolumn{2}{c}{\textbf{Posterior Reasoning Scores}} \\
\cmidrule(lr){2-3}
 & \textbf{Qwen-PRM} & \textbf{GPT-4o} \\
\midrule
LLaDA-8B-Base  & 0.31 & 0.36 \\
LLaDA-8B-Instruct & 0.38 & 0.43 \\
\bottomrule
\end{tabular}
\caption{\textbf{The \gls{mdlm} reasoning posterior yields high-quality traces for problems that the original instruct-tuned model fails to solve.} We perform posterior inference on the $1419$ training samples that LLaDA-8B-Instruct with greedy decoding fails to solve, and evaluate the resulting traces with two judges, \textsc{Qwen2.5-Math-PRM} \citep{zhang2025lessons} and \textsc{gpt-4}o \citep{hurst2024gpt}. Both judges rate $\sim40\%$ of the instruct-tuned reasoning chains as correct. Notably, even the posterior reasoning chains from the base model are rated as $>31\%$ correct.}\label{tab:judge_scores}
\end{table}

\begin{table}[t]
\centering
\small
\resizebox{\linewidth}{!}{%
\begin{tabular}{lccc}
\toprule
\textbf{Model} & \textbf{Post-training Data} & \textbf{GSM8K Test Acc.} & \textbf{Early-exit Speed Up} \\
\midrule
LLaDA 8B-Base (No template) & - & \textcolor{red}{13.9\%} &  \\

LLaDA 8B-Base (With template) & - & 51.2 \% &  \\
\quad Early exit $\gamma=1.0$ &  & 50.0 \% & $1.01\times$ \\

\midrule
Finetuned$^*$ LLaDA 8B-Base &
\textsc{gsm8}k $(\mbc, \mbr_\text{gold}, \mba_\text{gold})$ ($n=7473$) &
64.6 (\textcolor{teal}{$+13.4$}) \% &  \\
\quad Early exit $\gamma=0.3$ &  & 58.9 \% & $1.4\times$ \\
\quad Early exit $\gamma=0.4$ &  & 54.1 \% & $1.7\times$ \\
\midrule
Finetuned$^*$ LLaDA 8B-Base &
\textsc{gsm8}k Posterior $(\mbc, \mbr, \mba_\text{gold})$ ($n=7473$) &
66.1 (\textcolor{teal}{$+14.9$}) \% &  \\
\quad Early exit $\gamma=0.3$ &  & 66.2 \% & $1.2\times$ \\
\quad Early exit $\gamma=0.4$ &  & 64.7 \% & $1.5\times$ \\

\midrule
LLaDA-8B-Instruct & Misc. Instruction Data ($n=4.5$ million) & 75.96 \% & -- \\
\bottomrule
\end{tabular}%
}
\caption{\textbf{Fine-tuning the base model with posterior-generated data improves performance.}
Fine-tuning the LLaDA-8B-Base model on \textsc{gsm8}k human-written training data and posterior reasoning traces boosts accuracy to
$64.6\%$ and $66.1\%$, respectively. Finetuned$^*$ indicates LoRA \citep{hu2022lora} fine-tuning. Both fine-tuned models exhibit early exit capabilities. Notably, reasoning-as-infilling alone significantly improves base model performance from $13.9 \rightarrow 51.2\%$.}
\label{tab:finetune_gsm8k}
\end{table}

\subsection{Post-hoc reasoning} 

A key challenge for training better reasoning models is collecting high-quality reasoning traces for fine-tuning \citep{zelikman2022star}. We investigate whether the  \gls{mdlm}  posterior distribution, $p_\theta(\mbr \mid \mbc, \mba)$, can provide these traces, even when an \gls{mdlm} incorrectly solves the original task.

In this experiment, we generate samples from the LLaDA-8B-Instruct model with \gls{med} and any-order decoding. On the \textsc{gsm8}k training dataset \citep{lightman2023letsverifystepstep}, the model answers $1419$ out of $7473$ problems incorrectly. We use these $1419$ question-answer pairs to generate reasoning traces from both the instruct and the base model (without instruction tuning).

We evaluate these reasoning traces for correctness using \textsc{gpt4}o \citep{hurst2024gpt} and the Qwen2.5-Math-7B \textsc{prm} \citep{zhang2025lessons} (see \cref{appsec:gpt4_prompt} for the system instructions). As shown in \cref{tab:judge_scores}, both judges rate $\sim40\%$ of the instruct-tuned model's posterior reasoning traces as correct, even on problems the model originally fails. We observe that the traces judged correct by \textsc{gpt4}o contain accurate reasoning steps that correct the original model's errors (see \cref{appsec:posterior_traces} for examples). Notably, even the base model produces correct reasoning chains for $>30\%$ of these problems when conditioned on the answer. Posterior conditioning also outperforms STaR-style prefix-hint prompting \citep{zelikman2022star} with the same checkpoints for both instruct and base models (\cref{tab:hint_vs_posterior}).

\begin{figure}[t]
    \centering
    \begin{minipage}[c]{0.5\textwidth}
        \centering
        \includegraphics[width=\linewidth]{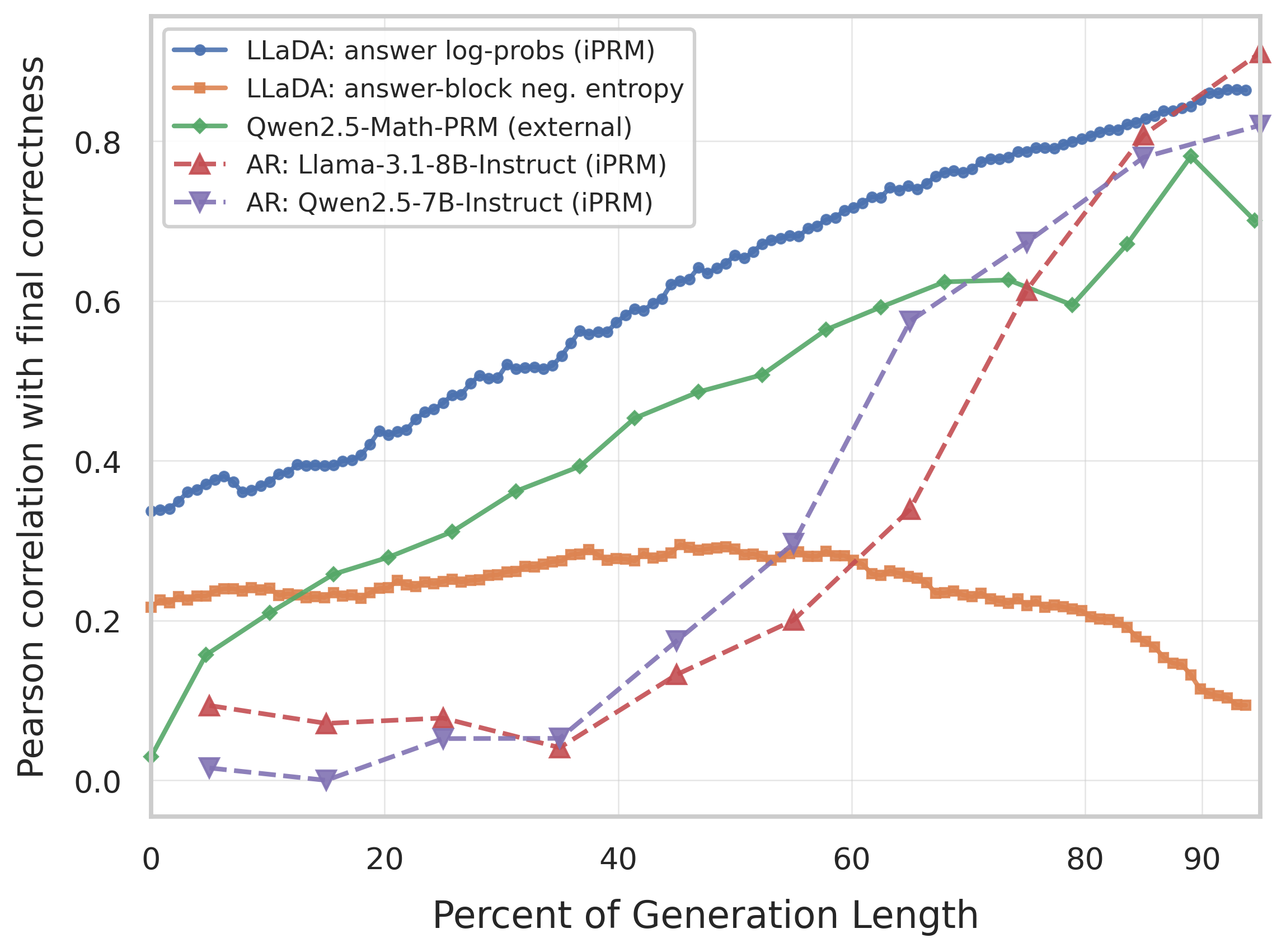}
    \end{minipage}%
    \hfill
    \begin{minipage}[c]{0.44\textwidth}
        \caption{\textbf{With answers, \glspl{mdlm} enable scoring their own reasoning process.}
        We score \textsc{gsm8}k reasoning traces, generated left-to-right with LLaDA, at intermediate steps using
        (a) gold answer log probabilities, (b) the answer block entropy bound, and (c) a 7B process reward model (\textsc{prm}) \citep{zhang2025lessonsdevelopingprocessreward}.
        Gold answer probability at intermediate steps is more predictive of final correctness than \textsc{prm} scores. We also compute an additional autoregressive baseline, where we compute gold answer likelihood scores at intermediate steps. The \textsc{ar} intermediate scores are significantly less correlated with final correctness (see \cref{appsec:ar_likelihood} for details).
        Even without gold labels, the \gls{mdlm} answer block entropy is weakly correlated with correctness.}\label{fig:correct_correlations}   
    \end{minipage}
    \vspace{-0.5cm}
\end{figure}

\paragraph{Posterior data can be used to improve base models.} Here, we examine the effectiveness of post-training on the post-hoc reasoning dataset generated on the full \textsc{gsm8}k training set with the \textit{base model}. We post-train the LLaDA-8B-Base model using LoRA \citep{hu2022lora} on a single Nvidia 40GB A100. 

In \cref{tab:finetune_gsm8k}, we observe that fine-tuning the model on the posterior generated base model data significantly improves performance ($+14.9\%$).
As a benchmark, fine-tuning on the \textsc{gsm8}k \textit{human annotated} reasoning traces produces comparable accuracy improvements on the test set. These results provide evidence that maximizing the log-likelihood $\log p_\theta(\mba, \mbr \mid \mbc)$ on the posterior reasoning traces improves accuracy on reasoning tasks. In turn, this indicates that when human reasoning traces are not available, \glspl{mdlm} have the capacity to bootstrap their own instruction fine-tuning datasets. We include additional training details in \cref{appsec:finetuning}.

Additionally, fine-tuning a model enhances early exiting capabilities. Using early exits can yield a $1.5\times$ acceleration with less than a $1.4$ percentage point drop in accuracy. In contrast, even at high $\gamma$ thresholds, early exits have limited effects when performing reasoning-as-infilling with the original base model.

\paragraph{Scoring Intermediate Reasoning Traces.} The reasoning-as-infilling framework also provides a mechanism for scoring reasoning traces at intermediate steps, which typically requires training a separate process reward model \citep{zhang2024rest}. Because reasoning-as-infilling maintains explicit answer block positions, the model's own conditional distributions over answer tokens can serve as an intermediate score as defined in \cref{eq:iprm}. 

Using LLaDA-8B-Instruct, we greedily sample solutions on the \textsc{gsm8}k test set (left-to-right, one token at a time) and compute the Pearson correlation between intermediate rewards and final output correctness. In \cref{fig:correct_correlations}, we compare several strategies for estimating the final correctness of partial reasoning traces (see \cref{appsec:posterior_filtering} for details). The answer log probabilities $\log p_\theta(a_j = a_j^* \mid \mbc, \mbr_{\textsc{un-masked}})$ correlate more strongly with final answer correctness at intermediate steps than scores from Qwen2.5-Math-PRM, a 7B process reward model \citep{zhang2025lessons}.

Additionally, we compare whether next-token prediction trained auto-regressive models can similarly score intermediate reasoning traces. In theory, auto-regressive models can also estimate the likelihood of a reference answer given a partial reasoning trace in a single forward pass, by appending \textit{The answer is \textbackslash boxed\{[ANSWER]} and computing answer likelihoods. In \cref{fig:correct_correlations}, we include results for Llama-3.1-8B-Instruct \citep{dubey2024llama3herdmodels} and Qwen2.5-7B-Instruct \citep{qwen2.5}. Notably, intermediate scores for these models are significantly less predictive of correctness at early and intermediate steps than both the $\phi_\textsc{iprm}$ and the dedicated process reward model. At early steps, intermediate auto-regressive likelihoods are less predictive of correctness than using label-free MDLM answer entropy.

These results suggest that \gls{mdlm} pre-training, combined with reasoning-as-infilling, opens up several post-training directions: low-quality reasoning chains could be terminated early, the reasoning process could be steered towards correct solutions, reflection tokens could be automatically inserted at reasoning failures, and dense intermediate feedback could be incorporated into fine-tuning objectives.

\section{Discussion}
Much of the current tooling around pre-training, post-training, and inference for text generation has been built around a key modeling choice: next-token prediction training. \glspl{mdlm} are an expressive class of models trained to in-fill masked sequences, requiring additional training and inference compute. In our work, we find that this additional compute has many uses, and enables \textit{rethinking} how these models are applied to reasoning tasks. Reasoning-as-infilling elicits useful properties of the \gls{mdlm} training objective, allowing the user to structure the reasoning process, accelerating inference, and providing new post-training capabilities. 

\paragraph{Limitations.} Our proposed post-training methods, both generating a reasoning trace conditioned on the answer and scoring reasoning traces at intermediate reasoning steps, require access to question-answer pairs. Therefore, they do not apply to post-training tasks where we do not have access to an answer. However, access to question-answer pairs is a standard assumption in reinforcement learning with verifiable rewards, for instance \textsc{grpo} \citep{deepseekai2025deepseekr1incentivizingreasoningcapability}.  Additionally, we limit our evaluations to tasks with short answers. Future work could explore early exit and post-hoc reasoning capabilities of diffusion models on tasks with long or open-ended answers.

\section*{Acknowledgments}
The authors would like to thank Subham Sahoo, Rico Angell, Chandhru Karthick, Mark Goldstein, Shenglong Wang, and Dinghuai Zhang for their valuable feedback.

This work was partly supported by the NIH/NHLBI Award R01HL148248, NSF Award 1922658 NRT-HDR: FUTURE Foundations, Translation, and Responsibility for Data Science, NSF CAREER Award 2145542, ONR N00014-23-1-2634, and Apple. Additional support was provided by a Fellowship from the Columbia Center of AI Technology. This work was also supported by IITP with a grant funded by the MSIT of the Republic of Korea in connection with the Global AI Frontier Lab International Collaborative Research. Zachary Horvitz is supported by a Google PhD Fellowship.

\section*{Large Language Model Use}
We used LLM assistants for proofreading, copyediting, and limited wording revisions. Additionally, we used AI coding tools for assistance with scripts and plot generation. All resulting outputs were reviewed by the authors.

\bibliography{text_diffusion_reasoning}
\bibliographystyle{colm2026_conference}

\appendix
\crefalias{section}{appendix}
\crefname{appendix}{appendix}{appendices}
\Crefname{appendix}{Appendix}{Appendices}

\section{Accelerated sampling with Multi-token Entropy Decoding}\label{appsec:med}

As \glspl{mdlm} learn the conditional distribution $p_\theta(x^j \mid \mbx_{\textsc{unmasked}})$ for all masked tokens, they support unmasking multiple tokens in parallel. However, decoding even two positions, $x^i$ and $x^j$ in parallel can result in samples that may not be likely under the \gls{mdlm} joint distribution $p_\theta(\mbx)$, as typically $p_\theta(x^i, x^j \mid \mbx_{\textsc{unmasked}}) \neq p_\theta(x^i \mid \mbx_{\textsc{unmasked}}) p_\theta(x^j \mid \mbx_{\textsc{unmasked}})$. 

However, for any set of positions $A \subseteq \textsc{mask-set} \subseteq \{1, \dots, L\}$, we can upper bound the \gls{kl} divergence between the joint distribution $p_\theta(\mbx^{A} \mid \mbx_\textsc{unmasked}, \mbc)$ and the factorized distribution $\prod_{i\in A} p_\theta( x^i \mid \mbx_\textsc{unmasked}, \mbc)$ with the sum of the entropies of the masked tokens: 
\begin{align}\label{eq:kl_bound}
    \kl{p_\theta(\mbx^{A} \mid \mbx_\textsc{unmasked}, \mbc) \Bigg| \prod_{i\in A} p_\theta( x^i \mid \mbx_\textsc{unmasked}, \mbc)} 
    \leq \sum_{i \in A} H(x^i \mid \mbx_\textsc{unmasked}, \mbc)
\end{align}
where $H$ is the entropy of the distribution $p_\theta(x^i \mid  \mbx_\textsc{unmasked}, \mbc)$. For a proof, see \cref{appsec:proofs}.

In this work, we additionally propose multi-token entropy decoding, which makes use of the entropies of the masked positions $x^j$ to decide whether to decode multiple positions in parallel. Given unmasked text $\mbx_{\textsc{unmasked}}$, a decoding threshold $\lambda$ and a maximum number of tokens to be decoded $k_\text{max}$, we propose the following definition of the set $A$ for selecting positions to un-mask. 
In \gls{med}, we sort the position entropies in an ascending order and decode positions that satisfy $H(x^i \mid \mbx_\textsc{unmasked}, \mbc) < \lambda$  and select $k_\text{max}$ such tokens. If no position has entropy lower than $\lambda$, we choose the position with the lowest entropy. Whenever the selected positions satisfy the entropy threshold, \cref{eq:kl_bound} is at most $\lambda |A| \leq \lambda k_\text{max}$. When no position satisfies the threshold, we instead decode the single minimum-entropy position as a fallback.

\citet{ben2025accelerated} previously proposed entropy-bound (\textsc{EB})-sampler, an adaptive multi-token decoder for
MDLMs, which controls the error incurred by multi-token decoding. \textsc{EB-sampler} adds multiple positions to unmask based on the difference between the sum of the positions added and
the maximum entropy until the difference exceeds a specified threshold $\gamma$. \citet{wu2025fast}
also propose accelerating inference with MDLMs using KV-caching and parallel decoding. Similar to
\citet{ben2025accelerated}, they propose an adaptive greedy strategy Block-wise Confidence-aware Parallel Decoding (\textsc{bpcd}), and decode tokens with
confidences above a fixed probability threshold. The method proposed in \citep{wu2025fast} is designed to accelerate argmax sampling from MDLMs.

\begin{figure*}[t]
    \centering
        \includegraphics[width=0.5\linewidth]{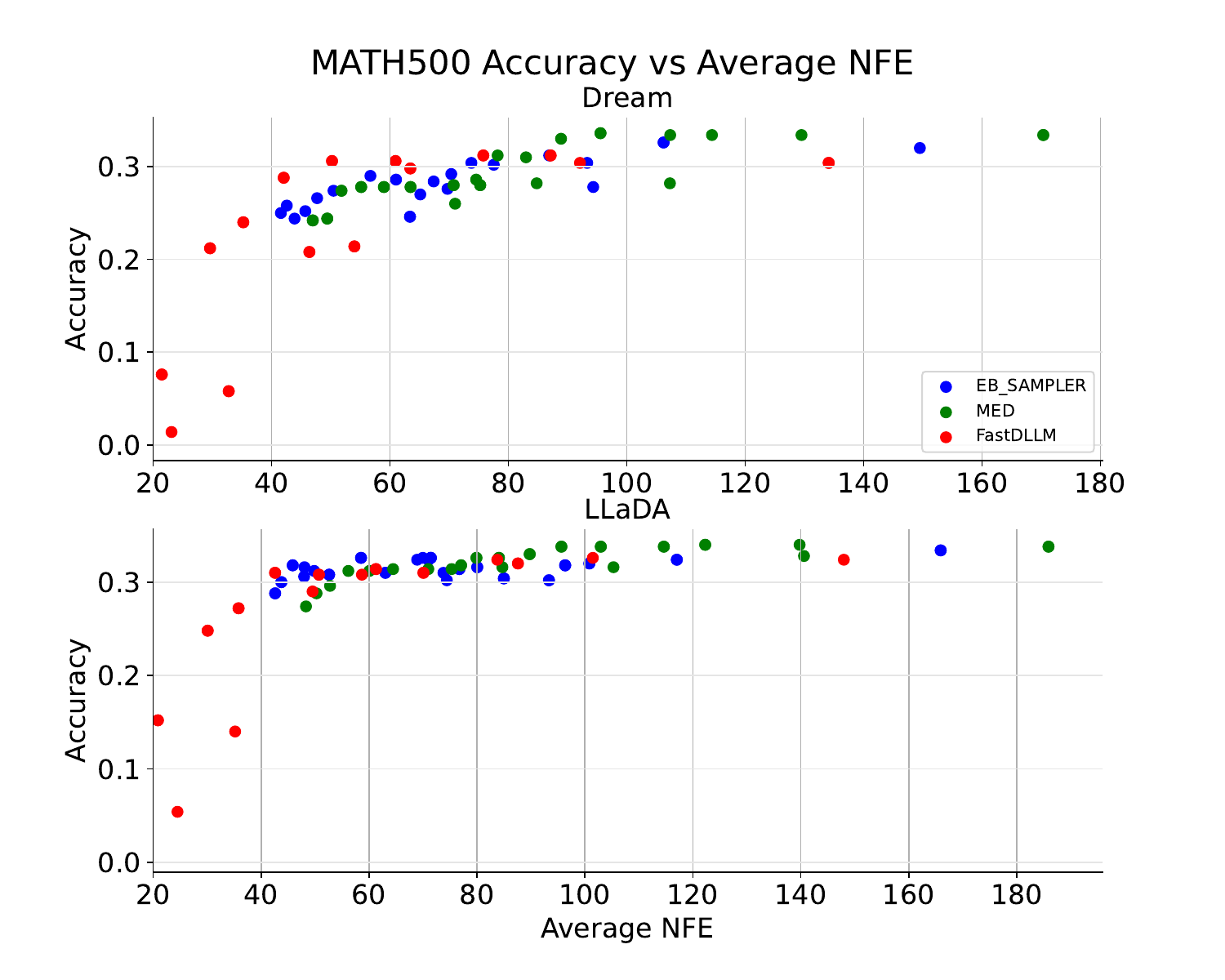}\includegraphics[width=0.5\linewidth]{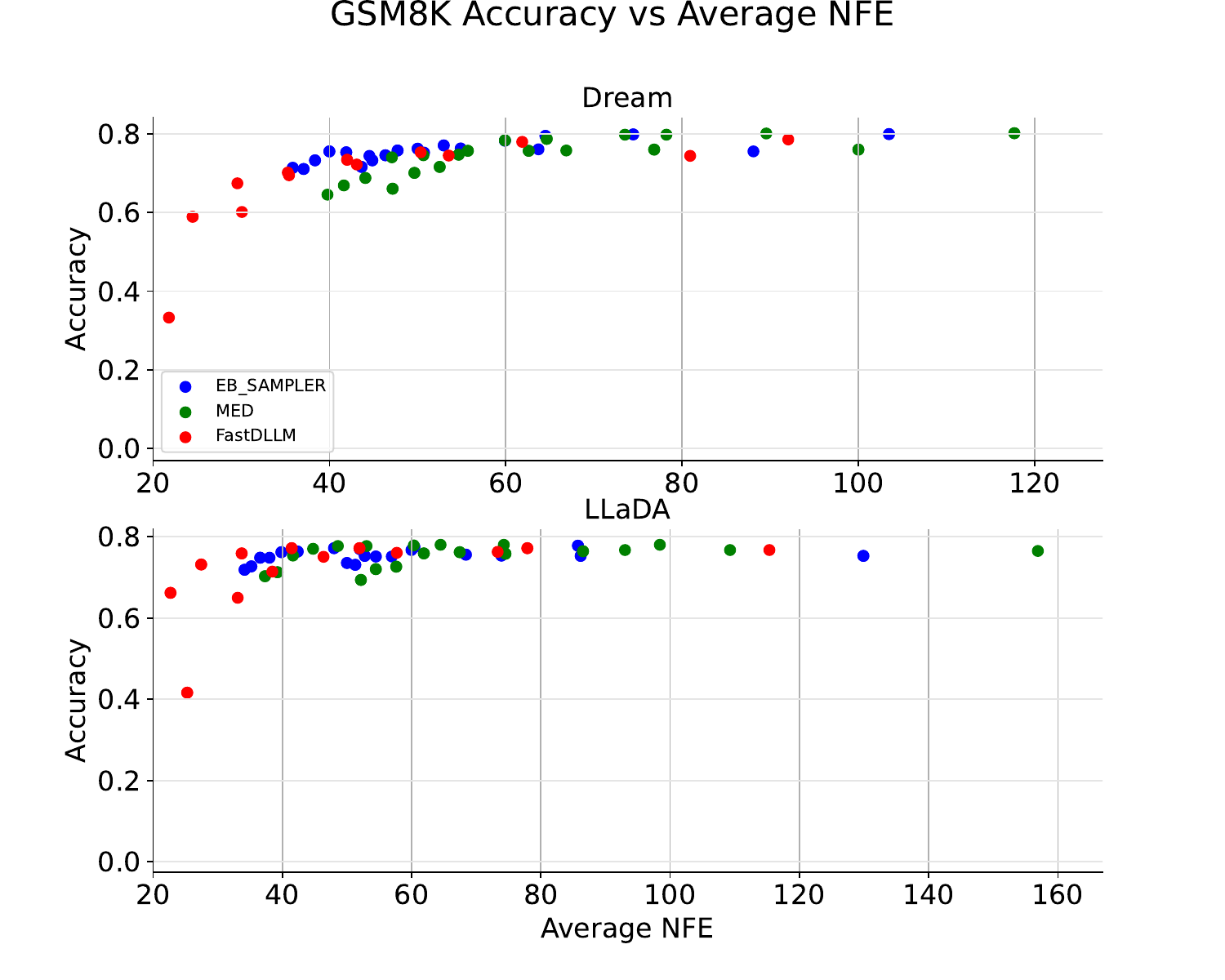}
        \caption{\textbf{Adaptive decoding can reduce function evaluations while retaining competitive accuracy.} In contrast to fixed token decoding, adaptive parallel decoding algorithms can preserve performance at selected operating points while reducing the number of function evaluations. Here, we plot results from \gls{med} alongside two recently proposed adaptive samplers, EB-Sampler \citep{ben2025accelerated} and Block-wise confidence-aware parallel decoding \citep{wu2025fast}. In this experiment, we generate sequences of length $L \in \{128, 256\}$ with varying thresholds for the three samplers. The best sampler depends on task and \textsc{nfe} budget. For instance, for a $2\times$ reduction in \textsc{nfe}s, \gls{med} yields the highest accuracies on \textsc{math500}, while producing comparable numbers to \textsc{eb}-sampler for \textsc{gsm8}k.}
        \label{fig:multi_token_kl}
\end{figure*}

In our sampler evaluations, we measure the task accuracy and the number of function evaluations (\textsc{nfe}s) for varying values of $k$ in the fixed token decoding scheme as well as varying values of $\lambda \in \{0.1, 0.2, 0.3, 0.4, 0.5, 0.6, 0.7, 0.8, 0.9\}$ in \gls{med}, \textsc{eb-sampler} \citep{ben2025accelerated}, and \textsc{bpcd} \citep{wu2025fast} with $k_{\text{max}}=32$ as the maximum number of tokens decoded in parallel. We fix a generation length $L \in \{128,256\}$ and a block size of $32$. In \cref{fig:multi_token_kl}, we compare the three adaptive decoding schemes with varying thresholds on the \textsc{gsm8k, math500} datasets. We observe that the best sampler depends on task and \textsc{nfe} budget. For instance, for a $2\times$ reduction in \textsc{nfe}s, \gls{med} yields the highest accuracies on \textsc{math500}, while producing comparable numbers to \textsc{eb}-sampler for \textsc{gsm8}k.

\begin{figure}[t]
    \centering
    \includegraphics[width=0.75\linewidth]{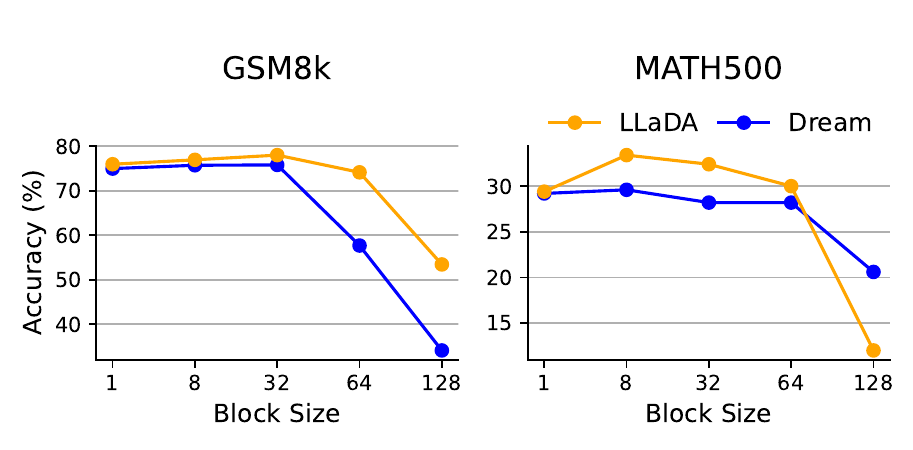}
    \caption{\textbf{Any-order decoding hurts performance on reasoning tasks.} When performing entropy decoding \citep{ye2025dream}, we observe that naive any-order sampling results in poor performance on all tasks. Left-to-right block decoding is required to make any-order sampling performant, and left-to-right sampling (block size $=1$) is a strong baseline. We also observe in \cref{tab:decoding_strategies} in \cref{appsec:observations} that performant block any-order configurations sample a large portion of tokens left-to-right. We consider sequences of length $128$.}
    \label{fig:decoding_strategies}
\end{figure}

\section{\texorpdfstring{\gls{mdlm}}{MDLM} Any-Order Sampling Behavior}\label{appsec:observations}

\begin{table}[h]
    \centering
    \small
    \resizebox{\linewidth}{!}{%
    \begin{tabular}{lcccccccc}
    \toprule
       \textbf{Decoding order}
       & \multicolumn{2}{c}{\textbf{GSM8K}}
       & \multicolumn{2}{c}{\textbf{Math500}}
       & \multicolumn{2}{c}{\textbf{HumanEval}}
       & \multicolumn{2}{c}{\textbf{Sudoku (4)}} \\
    \cmidrule(lr){2-3} \cmidrule(lr){4-5} \cmidrule(lr){6-7} \cmidrule(lr){8-9}
       & Llada & Dream 
       & Llada & Dream 
       & Llada & Dream 
       & Llada & Dream \\
    \midrule       
        Left-to-right   & 75.96  & 74.98  & 29.4 & 29.2  & 15.24  & \textbf{53.65}  & 36.13 & 17.28 \\
        Any-order decoding      & 53.44  & 34.11  & 12.0  & 20.6  & 10.97  & 32.92  & \textbf{47.64} & \textbf{61.26} \\
        \midrule
        Block any-order (8)        & 76.95  &  75.73  & \textbf{33.4}  & \textbf{29.6}  & \textbf{16.46}  & 51.82 & 38.74 & 44.50  \\
        Block any-order (32)       & \textbf{78.01}  & \textbf{75.81}  & 32.4  & 28.2 & 15.24 & 48.17  & \textbf{47.64}$^*$ & \textbf{61.26}$^*$ \\
        Block any-order (64)       & 74.14 & 57.69  & 30.0  & 28.2  & 14.63 & 47.56  & - & - \\
        \midrule
        Llama3.1-8B & \multicolumn{2}{c}{70.81} & \multicolumn{2}{c}{26.80} & \multicolumn{2}{c}{62.20} & \multicolumn{2}{c}{2.09} \\
    \bottomrule
    \end{tabular}%
    }
    \caption{\textbf{Left-to-right sampling is a competitive sampling algorithm for reasoning and coding.} When performing entropy decoding \citep{ye2025dream}, we observe that full any-order sampling results in poor performance on all tasks but Sudoku. Left-to-right block decoding is required to make any-order sampling performant on the non-Sudoku tasks, and on those tasks left-to-right sampling (block size $=1$) is within a few percentage points of the best configuration. We also observe in \cref{appsec:observations} that performant block any-order configurations sample a large portion of tokens left-to-right.$^*$For Sudoku, we consider sequences of length 32, otherwise we use a sequence length of $128$.} 
    \label{tab:decoding_strategies}
\end{table}

We study the effects of greedy any-order entropy decoding \citep{dream2025} for LLaDA \citep{nie2025largelanguagediffusionmodels} and Dream \citep{dream2025}, as well as any-order with different \textbf{block lengths} \citep{sahoo2024simple,arriola2025blockdiffusioninterpolatingautoregressive, nie2025largelanguagediffusionmodels, dream2025}. The block length is the contiguous region of consecutive positions considered by the sampling algorithm, where the model can decode in any order. Blocks are unmasked left-to-right.

We include our results in \cref{tab:decoding_strategies}. On Sudoku, any-order sampling substantially improves performance.\footnote{Of note, on Sudoku, diffusion models with auto-regressive sampling substantially outperform Llama 8B. This may reflect benefits of the \gls{mdlm} training objective.} However, for the remaining datasets, left-to-right sampling with a block length of $1$ is a competitive approach. In some cases (e.g. for Dream on \textsc{HumanEval} \citep{chen2021evaluating}), left-to-right block length $1$ sampling is the most performant configuration. Additionally, purely any-order decoding (i.e. when the block size $=$ generation length) often leads to a large drop in performance outside Sudoku. 

\textbf{In what order are tokens decoded?}  In \cref{tab:gsm8k_behavior} and \cref{tab:humaneval_behavior}, we analyze the behavior of these different configurations on a portion of \textsc{gsm8k} and \textsc{HumanEval}. We compute the fraction of non-EOS tokens decoded from the leftmost masked position, the average distance from the leftmost position, and the total number of non-EOS tokens. For \textsc{gsm8k}, we also include the average step at which the answer appears in the decoded sequence.

Notably, top performing block-length configurations often behave very autoregressively. On \textsc{gsm8k}, when the block size is $32$, both LLaDA and Dream sample the leftmost unmasked position approximately $50\%$ of the time. Additionally, the average distance of the unmasked position from the left-most mask is approximately $3$ tokens. \citet{gong2025diffucoder} similarly observe the left-to-right sampling behavior of Dream for coding.

\textbf{Why does block sampling improve performance?} We find that purely any-order decoding from current \glspl{mdlm} results in less auto-regressive generation. On \textsc{gsm8k}, it also produces fewer non-EOS tokens and much earlier answers, leaving more of the allocated generation length unused. Reviewing samples from any-order  decoding, we observe two specific pathological behaviors: 1) Models first greedily decoding low entropy \textit{end-of-text} tokens, leading to shorter or empty texts that do not fully utilize the assigned tokens, and 2) \textit{decoding only an answer, or decoding answers first, {before} reasoning chains}. We include an example of the latter behavior below.

\lstset{
  basicstyle=\ttfamily\small,
  breaklines=true,
  breakatwhitespace=true,
  columns=fullflexible,
  keepspaces=true,
  escapeinside={(*}{*)}
}

\newtcolorbox{examplebox}[1][]{
  breakable,
  colback=gray!5,
  colframe=black!40,
  title=#1,
  fonttitle=\bfseries,
  coltitle=black,
  left=1em,
  right=1em,
  top=0.5em,
  bottom=0.5em
}

\begin{examplebox}[GSM8K Example: Any-Order Entropy Decoding (LLaDA)]
\textbf{Question.} Kylar went to the store to buy glasses for his new apartment. One glass costs \$5, but every second glass costs only 60\% of the price. Kylar wants to buy 16 glasses. How much does he need to pay for them?

\vspace{6pt}
\textbf{Gold answer.} \textbf{64}

\vspace{6pt}
\textbf{Final generation (incorrect).}
\begin{lstlisting}
Kylar wants to buy 16 glasses, so he needs to pay for 16/2 = 8 glasses at the full price.
The cost for these 8 glasses is 6 * $5 = $30.
The remaining 8 glasses are at the discounted price, which is $5 * 60% = $3.6.
The cost for these 8 glasses is 5 * $3.6 = $18.
In total, Kylar needs to pay $30 + $18 = 48.
Solution: 48
\end{lstlisting}

\vspace{6pt}
\textbf{Decoding history (compressed).}
\begin{lstlisting}
Step 2:
<|eot_id|>

Step 5:
<...masks...>.<...masks...><|eot_id|>

Step 10:
<...masks...>8.
Solution: (*\textcolor{red}{48}*)<|eot_id|>

Step 20:
<...masks...> $30 + $18 = (*\textcolor{red}{48}*).
Solution: (*\textcolor{red}{48}*)<|eot_id|>

Step 40:
Kylar<...masks...> = $18.
In total, Kylar needs to pay $30 + $18 = (*\textcolor{red}{48}*).
Solution: (*\textcolor{red}{48}*)<|eot_id|>

Step 127 (final output):
Kylar wants to buy 16 glasses, so he needs to pay for 16/2 = 8 glasses at the full price.
The cost for these 8 glasses is 6 * $5 = $30.
The remaining 8 glasses are at the discounted price, which is $5 * 60% = $3.6.
The cost for these 8 glasses is 5 * $3.6 = $18.
In total, Kylar needs to pay $30 + $18 = (*\textcolor{red}{48}*).
Solution: (*\textcolor{red}{48}*)<|eot_id|>
\end{lstlisting}

\vspace{6pt}
\textbf{Observation.} Under any-order entropy decoding, LLaDA commits to the incorrect final answer early (by step 10), and subsequently fills in a post-hoc reasoning trace that is consistent with this answer.
\end{examplebox}

\begin{table}[h]
    \centering
    \small
    \resizebox{\linewidth}{!}{%
    \begin{tabular}{l c ccccc}
        \toprule
        \textbf{Config} & \textbf{Model} & \textbf{Acc.} & \textbf{\% Leftmost} & \textbf{Dist. Left} & \textbf{Non-EOS Tokens} & \textbf{Answer Step} \\
        \midrule
        \multirow{2}{*}{Block(1)}     
            & Dream  & 76.4\% & 100.0\% & 0.0 & 105.1 & 78.1 \\
            & LLaDA  & 79.0\% & 100.0\% & 0.0 & 115.3 & 84.3 \\
            \midrule
        \multirow{2}{*}{Block(32)}  
            & Dream  & 77.6\% & 52.1\%  & 2.9 & 103.1 & 76.3 \\
            & LLaDA  & 76.8\% & 47.1\%  & 3.3 & 112.3 & 82.8 \\
                        \midrule

        \multirow{2}{*}{AO(128)} 
            & Dream  & 34.2\% & 73.1\%  & 6.5 & 24.3  & 18.7 \\
            & LLaDA  & 53.4\% & 40.8\%  & 20.2 & 75.0  & 16.9 \\
        \bottomrule
    \end{tabular}%
    }
        \caption{\textbf{Decoding behavior, GSM8K} We evaluate the autoregressiveness of different sampling configurations by measuring the percent of non-EOS tokens decoded from the leftmost position, the average distance of these positions from left, the total number of non-EOS tokens, and at what timestep the answer is decoded. We consider generation lengths of $128$ on a portion of GSM8K ($n=500$)}
        \label{tab:gsm8k_behavior}
\end{table}

\begin{table}[h]
    \centering
    \small
    \resizebox{\linewidth}{!}{%
    \begin{tabular}{l c cccc}
        \toprule
        \textbf{Config} & \textbf{Model} & \textbf{Acc.} & \textbf{\% Leftmost} & \textbf{Dist. Left} & \textbf{Non-EOS Tokens} \\
        \midrule
        \multirow{2}{*}{Block(1)}     
            & Dream  & 53.7\% & 100.0\% & 0.0 & 95.0  \\
            & LLaDA  & 11.0\% & 100.0\% & 0.0 & 119.8 \\
                        \midrule

        \multirow{2}{*}{Block(32)}  
            & Dream  & 48.2\% & 43.1\%  & 3.8 & 96.8  \\
            & LLaDA  & 15.2\% & 44.7\%  & 4.1 & 119.5 \\
            \midrule
        \multirow{2}{*}{AO(128)} 
            & Dream  & 32.9\% & 44.5\%  & 6.8 & 56.4  \\
            & LLaDA  & 15.2\% & 29.7\%  & 19.7 & 123.8 \\
        \bottomrule
    \end{tabular}%
    }
              \caption{\textbf{Decoding behavior, HumanEval} Similar to \cref{tab:gsm8k_behavior}, we measure autoregressiveness for generation lengths of $128$ on HumanEval ($n=164$).}
            \label{tab:humaneval_behavior}
\end{table}

\section{Evaluating Reasoning Trace Correctness}\label{appsec:gpt4_prompt}

We evaluate reasoning trace correctness with \textsc{gpt4}o \citep{hurst2024gpt}. We use the following template to evaluate the reasoning chain:

\begin{verbatim}
SYSTEM_INSTRUCTIONS = """
You are a strict grader.
You are given a question and a model's response.

Return a answer of either:
- 1 if both the model's final answer AND reasoning chain are correct.
- 0 otherwise.

Your final answer should be the digit 0 or 1.
"""


PROMPT_TEMPLATE = """Grade the submission.

[Question]
{question}

[Model Response]
{reasoning}

[Reference Answer]
{reference_answer}
"""
\end{verbatim}

We use $512$ tokens for all judge models, and default hyperparameters.

\section{Reasoning Trace Examples}\label{appsec:posterior_traces}

Below, we include reasoning traces from the model $p_\theta(\mba \mid \mbr, \mbc) p_\theta(\mbr \mid \mbc)$ and the posterior traces $p_\theta(\mbr \mid \mba, \mbc)$. We include samples where LLaDA originally fails to correctly answer the question. In two cases, the posterior successfully provides an alternative correct reasoning trace, and in two it fails to do so.

\subsection{Correct Posterior Traces}

\begin{examplebox}[Example 1]
\textbf{Question:} Ms. Warren ran at 6 mph for 20 minutes. After the run, she walked at 2 mph for 30 minutes. How many miles did she run and walk in total? \\[6pt]

\textbf{Incorrect LLaDA output:}
\begin{lstlisting}
Ms. Warren ran at 6 mph for 20 minutes, which is $\frac{20}{60} = \frac{1}{3}$ hours. So, she ran $6 \times \frac{1}{3} = 2$ miles.

Ms. Warren walked at 2 mph for 30 minutes, which is $\frac{30}{60} = \frac{1}{2}$ hours. So, she walked $2 \times \frac{1}{2} = (*\textcolor{red}{2 miles}*) as well.

The answer is \boxed{(*\textcolor{red}{4}*)} miles.
\end{lstlisting}

\textbf{Correct LLaDA posterior:}
\begin{lstlisting}
Ms. Warren ran at 6 mph for 20 minutes, which is 20/60 = 1/3 of an hour.
So, she ran 6 mph * 1/3 hour = 2 miles.
After the run, she walked at 2 mph for 30 minutes, which is 30/60 = 1/2 of an hour.
So, she walked 2 mph * 1/2 hour = 1 mile.
In total, Ms. Warren ran and walked 2 miles + 1 mile = 3 miles.
The answer is \boxed{3}
\end{lstlisting}
\end{examplebox}

\begin{examplebox}[Example 1]
\textbf{Question:} Each week Jaime saves \$50. Every two weeks she spends \$46 of her savings on a nice lunch with her mum. How long will it take her to save \$135? \\[6pt]

\textbf{Incorrect LLaDA output:}
\begin{lstlisting}
Jaime saves $50 each week, so in two weeks she saves $50 x 2 = $100.
Every two weeks she spends $46, so in two weeks she saves $100 - $46 = $54.
To save $135, she needs to save $135 - $54 = $81 more.
Since she saves $54 every two weeks, it will take her $81 / $54 = (*\textcolor{red}{3}*) more two weeks to save $135.
The answer is \boxed{3}+2 = (*\textcolor{red}{6 weeks}*).
\end{lstlisting}

\textbf{Correct LLaDA posterior:}
\begin{lstlisting}
Jaime saves $50 each week, so in two weeks she saves $50 x 2 = $100.
However, she spends $46 every two weeks, so her net savings every two weeks is $100 - $46 = $54.
To find out how long it will take her to save $135, we divide $135 by her net savings every two weeks: $135 / $54 = 2.5.
Therefore, it will take her 2.5 two weeks to save $135.
The answer is \boxed{5}
\end{lstlisting}
\end{examplebox}

\subsection{Incorrect Posterior Traces}

\begin{examplebox}[Example 1]
\textbf{Question:} Tommy is making 12 loaves of bread. He needs 4 pounds of flour per loaf. A 10-pound bag of flour costs \$10 and a 12-pound bag costs \$13. When he is done making his bread, he has no use for flour and so he will throw away whatever is left. How much does he spend on flour if he buys the cheapest flour to get enough? \\[6pt]

\textbf{Incorrect LLaDA output:}
\begin{lstlisting}
Tommy needs 12 loaves of bread, and each loaf requires 4 pounds of flour. 
Therefore, he needs a total of 12 * 4 = 48 pounds of flour.

The cheapest option is to buy 12-pound bags of flour. 
Since Tommy needs 48 pounds, he will need 48/12 = 4 bags. 
Each bag costs $13, so the total cost for 4 bags is 4 * 13 = $52.

The answer is \boxed{(*\textcolor{red}{52}*)}.
\end{lstlisting}

\textbf{Incorrect LLaDA posterior:}
\begin{lstlisting}
Tommy needs 12 loaves of bread, so he needs a total of 12 * 4 = 48 pounds of flour.
If he buys 10-pound bags, he will need 48/10 = (*\textcolor{red}{4 bags}*), 
which will cost 4 * 10 = (*\textcolor{red}{40}*).
If he buys 12-pound bags, he will need 48/12 = 4 bags, which will cost 4 * 13 = $52.
Therefore, Tommy should buy 10-pound bags to save money.
The answer is \boxed{50}.
\end{lstlisting}
\end{examplebox}

\begin{examplebox}[Example 2]
\textbf{Question:} Thomas started saving for a car almost 2 years ago. For the first year, his weekly allowance was \$50. In the second year, he got a job that pays \$9 an hour at a coffee shop and worked 30 hours a week, so his parents discontinued his allowance. If the car he wants to buy is \$15{,}000 and he spends \$35 a week on himself, how much more money does Thomas need to buy the car by the end of the 2 years? \\[6pt]

\textbf{Incorrect LLaDA output:}
\begin{lstlisting}
In the first year, Thomas earned $50 x 52 = $2600.
In the second year, Thomas earned $9 x 30 x 52 = (*\textcolor{red}{7980}*).
Thomas saved $2600 + $7980 = (*\textcolor{red}{10580}*).
Thomas spent $35 x (*\textcolor{red}{52}*) = (*\textcolor{red}{1820}*).
Thomas still needs $15000 - 10580 - 1820 = (*\textcolor{red}{2600}*) more.
The answer is \boxed{(*\textcolor{red}{2600}*)}.
\end{lstlisting}

\textbf{Incorrect LLaDA posterior:}
\begin{lstlisting}
In the first year, Thomas earned $50 x 52 = $2600.
In the second year, Thomas earned $9 x 30 x 52 = (*\textcolor{red}{8760}*).
Thomas spent $(*\textcolor{red}{35 x 52 x 52}*) = (*\textcolor{red}{8360}*).
In total, Thomas saved $2600 + $8760 - $8360 = (*\textcolor{red}{3000}*).
To buy the car, Thomas still needs $15000 - 3000 = $2000.
The answer is \boxed{2000}
\end{lstlisting}
\end{examplebox}

\begin{table}[t]
\centering \small
\begin{tabular}{llc}
\toprule
\textbf{Model} & \textbf{Generation method} & \textbf{GPT-4o score} $\uparrow$ \\
\midrule
LLaDA-8B-Instruct & STaR-style hint      & 37.0\% \\
LLaDA-8B-Instruct & Posterior reasoning  & \textbf{44.1\%} \\
\midrule
LLaDA-8B-Base     & STaR-style hint      & 13.5\% \\
LLaDA-8B-Base     & Posterior reasoning  & \textbf{33.9\%} \\
\bottomrule
\end{tabular}
\caption{\textbf{Posterior reasoning vs.\ prefix-hint prompting, same checkpoints.} On a $1000$-question subset of the \textsc{gsm8}k \emph{training} questions that LLaDA-8B-Instruct fails to solve, posterior conditioning on the answer block yields traces judged correct more often than STaR-style hint prompting \citep{zelikman2022star} with the same model, with the largest gap for the base model.}
\label{tab:hint_vs_posterior}
\end{table}

\section{Scoring Posterior Reasoning Chains with MDLMs}\label{appsec:posterior_filtering}

Given that posterior sampling yields both high-quality and low-quality reasoning chains, a natural question is: \textit{Can we filter these chains without an external model?}

\begin{figure}[t]
    \centering
    \includegraphics[width=0.9\linewidth]{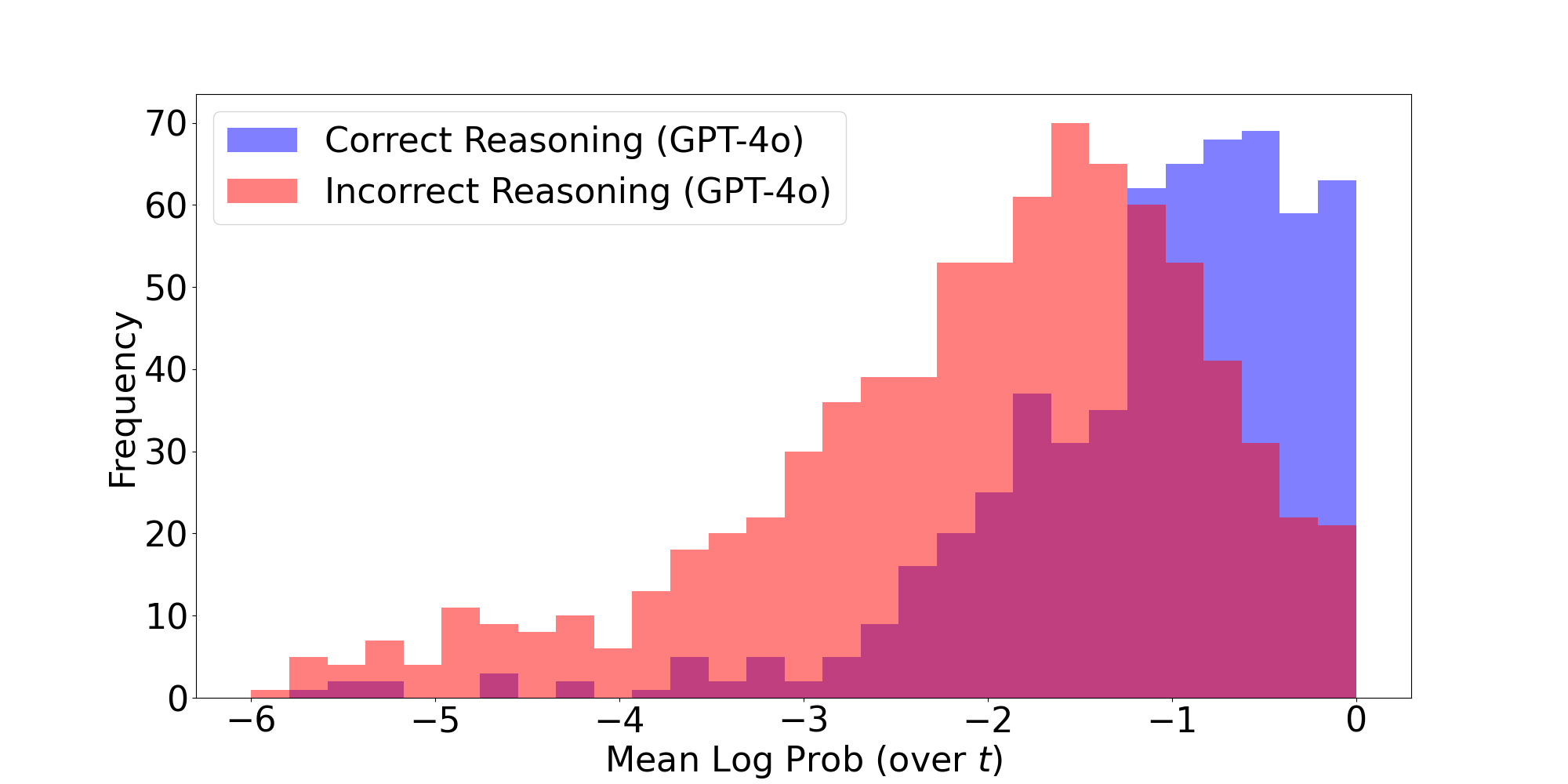}
    \caption{\textbf{Distribution of answer block log probability scores for posterior samples.} To score the $1419$ posterior reasoning traces, we iteratively unmask each chain left-to-right, and compute the average answer block log probabilities across all timesteps. Posterior chains rated as correct by \textsc{GPT4}-o tend to have higher scores.}
    \label{fig:posterior_logprob_dist}
\end{figure}

\begin{figure}[t]
    \centering
    \includegraphics[width=0.6\linewidth]{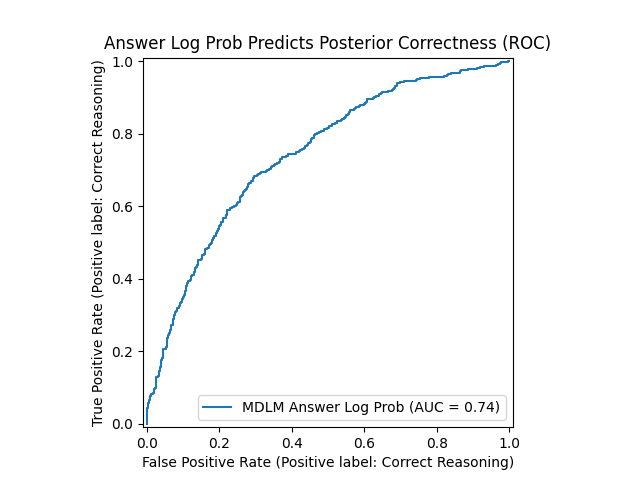}
    \caption{\textbf{Answer block log probability scores predict posterior reasoning trace quality.} The \gls{mdlm} average answer block log probabilities can simply be thresholded to provide a classifier for predicting \textsc{gpt4}-o reasoning chain judgments. This classifier provides a potential method for filtering low-quality posterior samples (e.g. before fine-tuning).}
    \label{fig:posterior_logprob_roc}
\end{figure}

We find that answer block log probabilities computed with reasoning-as-infilling can be used to filter these reasoning chains, and identify traces that \textsc{gpt4}-o rates as correct. To do this, we iteratively unmask each generated posterior chain, left-to-right, and average the answer log probabilities over all time-steps. For LLaDA, we observe that correct reasoning traces correspond to higher average scores, see \cref{fig:posterior_logprob_dist}. We also find that thresholding these answer-block log-probability scores yields AUC=$0.74$, see \cref{fig:posterior_logprob_roc}.

\section{Early Exit Samples}\label{appsec:exits}
Below we demonstrate outputs from LLaDA on the \textsc{gsm8}k dataset. For a sample question from the test set, we generate a sequence of length 256 with the following settings:
\begin{itemize}
    \item Sampling from the model, without using the reasoning template or early exits.
    \item Reasoning template with no exits. 
    \item Reasoning template with varying exit thresholds.
    \item Reasoning template with the model producing the answer with no reasoning tokens decoded.
\end{itemize}

\begin{examplebox}[Example 1]
\textbf{Question:} Janet's ducks lay 16 eggs per day. She eats three for breakfast every morning and bakes muffins for her friends every day with four. She sells the remainder at the farmers' market daily for \$2 per fresh duck egg. How much in dollars does she make every day at the farmers' market? \\[6pt]

\textbf{LLaDA (No reasoning template, No Early Exits):}
\begin{lstlisting}

To determine how much Janet makes every day at the farmers' market, we need to follow these steps:
1. Calculate the total number of eggs laid by Janet's ducks per day.
2. Subtract the number of eggs Janet eats and bakes for her friends from the total number of eggs laid.
3. Determine the number of eggs she sells at the farmers' market.
4. Calculate the total revenue from selling the eggs.
Step 1: Janet's ducks lay 16 eggs per day.
Step 2: Janet eats 3 eggs for breakfast every morning and bakes 4 eggs for her friends every day. Therefore, she eats and bakes a total of \(3 + 4 = 7\) eggs per day.
Step 3: The number of eggs she sells at the farmers' market is \(16 - 7 = 9\) eggs per day.
Step 4: Janet sells each egg for $2. Therefore, the total revenue from selling 9 eggs per day is \(9 \times 2 = 18\) dollars.
So, the amount Janet makes every day at the farmers' market is \(\boxed{18}\).
\end{lstlisting}

\textbf{LLaDA output (Reasoning template, No Early Exit):}
\begin{lstlisting}


To determine how much Janet makes every day at the farmers' market, we need to follow these steps:
Step 1: Calculate the total number of eggs laid by Janet's ducks per day.
Janet's ducks lay 16 eggs per day.
Step 2: Calculate the number of eggs Janet eats for breakfast per day.
Janet eats 3 eggs for breakfast every day.
Step 3: Calculate the number of eggs Janet bakes for her friends per day.
Janet bakes 4 eggs for her friends every day.
Step 4: Calculate the number of eggs left after Janet eats some and bakes some.
Total eggs laid - Eggs eaten for breakfast - Eggs bakes for friends = Eggs left
16 - 3 - 4 = 9
So, Janet has 9 eggs left to sell.
Step 5: Calculate the revenue from selling the eggs at the farmers' market.
Janet sells each egg for $2.
Eggs left * Price per egg = Total revenue
9 * 2 = 18
Therefore, Janet makes $18 every day at the farmers' market.
The answer is \boxed{18}.
\end{lstlisting}

\textbf{LLaDA output (Reasoning template, Early Exit($\gamma=0.5$)):}
\begin{lstlisting}
To determine how much Janet makes every day at the farmers' market, we need to follow these steps:
Step 1: Calculate the total number of eggs laid by Janet's ducks per day.
<|mdm_mask|>...<|mdm_mask|> per<|mdm_mask|>...<|mdm_mask|> The answer is \boxed{18}.
\end{lstlisting}

\textbf{LLaDA output (Reasoning template, Early Exit($\gamma=0.7$)):}
\begin{lstlisting}
To determine<|mdm_mask|>...<|mdm_mask|> The answer is \boxed{18}.
\end{lstlisting}

\textbf{LLaDA output (Forced Answer):}
\begin{lstlisting}
<|mdm_mask|>...<|mdm_mask|>. The answer is \boxed{(*\textcolor{red}{14}*)}.
\end{lstlisting}
\end{examplebox}

\section{Early-Exit Baseline Comparison and Threshold Sensitivity}\label{appsec:prophet}

We compare entropy-based early exits with Prophet \citep{li2025diffusion} on \textsc{gsm8}k and CommonsenseQA \citep{talmor2019commonsenseqa}. We use a generation length of $256$ with block size $32$, matching the early-exit setup in the main paper, and run Prophet using the configuration provided in the original paper. 

The results are shown in \cref{tab:prophet}. On CommonsenseQA, entropy-based exits are both more accurate and faster than Prophet at every evaluated threshold. On \textsc{gsm8}k, entropy exit matches Prophet at a comparable \textsc{nfe} budget ($+0.5$ percentage points at ${\sim}1.55\times$) and offers a smooth accuracy and speed tradeoff at lower thresholds.

\begin{table}[t]
\centering \small
\begin{tabular}{llccc}
\toprule
\textbf{Dataset} & \textbf{Method} & \textbf{Acc. (\%)} $\uparrow$ & \textbf{Avg. NFE} $\downarrow$ & \textbf{Speedup} $\uparrow$ \\
\midrule
\multirow{5}{*}{CommonsenseQA}
 & No exit                      & 72.3 & 256.0 & 1.00$\times$ \\
 & Prophet \citep{li2025diffusion} & 72.2 & 251.9 & 1.02$\times$ \\
 & Entropy exit, $\gamma{=}1.0$ & 74.7 & 171.3 & 1.49$\times$ \\
 & Entropy exit, $\gamma{=}1.2$ & \textbf{76.2} & 137.6 & 1.86$\times$ \\
 & Entropy exit, $\gamma{=}1.4$ & 75.7 & 102.7 & \textbf{2.49}$\times$ \\
\midrule
\multirow{5}{*}{GSM8K}
 & No exit                       & 78.8 & 256.0 & 1.00$\times$ \\
 & Prophet \citep{li2025diffusion} & 74.3 & 164.5 & 1.56$\times$ \\
 & Entropy exit, $\gamma{=}0.05$ & 78.5 & 225.6 & 1.13$\times$ \\
 & Entropy exit, $\gamma{=}0.1$  & 78.0 & 198.5 & 1.29$\times$ \\
 & Entropy exit, $\gamma{=}0.2$  & 74.8 & 164.8 & 1.55$\times$ \\
\bottomrule
\end{tabular}
\caption{\textbf{Comparison with Prophet and early-exit threshold sensitivity.} We evaluate Prophet \citep{li2025diffusion} using its published configuration. On CommonsenseQA, entropy-based exits are more accurate and faster at every evaluated threshold. On \textsc{gsm8}k, entropy exit matches Prophet at a comparable \textsc{nfe} budget and offers a smooth accuracy--speed tradeoff at lower thresholds.}
\label{tab:prophet}
\end{table}

\section{Autoregressive Answer-Likelihood Baselines}\label{appsec:ar_likelihood}

In this section, we provide details for the autoregressive answer-likelihood baselines reported in \cref{fig:correct_correlations}.

For each problem in the \textsc{gsm8}k test set ($1319$ problems), each autoregressive model (Llama-3.1-8B-Instruct and Qwen2.5-7B-Instruct) generates its own reasoning trace greedily (temperature $0$, one sample per problem) with a maximum budget of $128$ tokens. Every $5$ generated tokens, we perform a single additional forward pass in which the answer delimiter \textit{`` The answer is \textbackslash boxed\{"} is appended to the partial trace, and we compute
\begin{align}\nonumber
\phi_{\textsc{ar}}(\mbr_{\le t}) := \sum_{j} \log p(a^*_j \mid \mbc, \mbr_{\le t}, \text{delimiter}),
\end{align}
the sum of the log-probabilities of the reference answer tokens. Checkpoints are binned into ten equal fractions of each sample's total generation length, and within each bin we compute the Pearson correlation between $\phi_{\textsc{ar}}$ and the model's own final answer correctness. The full results are shown in \cref{fig:correct_correlations}.

We restrict the analysis to generations that terminate within the $128$-token budget (Llama: $703/1319$, Qwen: $578/1319$). This restriction is required to match the $128$-token generation budget used for the \gls{mdlm} curves in \cref{fig:correct_correlations}. Even in this setting, intermediate answer likelihoods for both models are substantially less predictive of final correctness than the \textsc{mdlm}-defined intermediate score, $\phi_{\textsc{iPRM}}$, through most of the generation.

\section{Proofs}\label{appsec:proofs}

\paragraph{\gls{med} \gls{kl} upper-bound.}
Here we prove that for any set $A \subset \{1, \dots\} / \textsc{un-masked}$, the following upper bound holds:
\begin{align}
    \kl{p_\theta(\mbx^{A} \mid \mbx_\textsc{un-masked}, \mbc) \Bigg| \prod_{i\in A} p_\theta( x^i \mid \mbx_\textsc{un-masked}, \mbc)} 
    \leq \sum_{i \in A} H(x^i \mid \mbx_\textsc{un-masked}, \mbc)
\end{align}
Note that:
\begin{align}\nonumber
   & \kl{p_\theta(\mbx^{A}  \mid \mbx_\textsc{un-masked}, \mbc) \Bigg| \prod_{i\in A} p_\theta( x^i \mid \mbx_\textsc{un-masked}, \mbc)}  \\
    & \qquad = \mbE_{p_\theta(\mbx^{A} \mid \mbx_\textsc{un-masked}, \mbc) } \left[ \log p_\theta(\mbx^{A} \mid \mbx_\textsc{un-masked}, \mbc)  - \log \prod_{i\in A} p_\theta(x^i \mid \mbx_\textsc{un-masked}, \mbc) \right]   \\ 
    & \qquad = \mbE_{p_\theta(\mbx^{A} \mid \mbx_\textsc{un-masked}, \mbc) } \left[ \log p_\theta(\mbx^{A} \mid \mbx_\textsc{un-masked}, \mbc)  - \sum_{i\in A} \log p_\theta(x^i \mid \mbx_\textsc{un-masked}, \mbc) \right]   \\   \label{eq:kl_entropy}  
    & \qquad = -H(\mbx^{A} \mid \mbx_\textsc{un-masked}, \mbc) + \sum_{i \in A} H(\mbx^i \mid \mbx_\textsc{un-masked}, \mbc) 
\end{align}
Now, since the entropy for discrete random variables is nonnegative, $H(\mbx^{A} \mid \mbx_\textsc{un-masked}, \mbc) \geq 0$, which implies:
\begin{align}
    -H(\mbx^{A} \mid \mbx_\textsc{un-masked}, \mbc) + \sum_{i \in A} H(\mbx^i \mid \mbx_\textsc{un-masked}, \mbc) \leq \sum_{i \in A} H(\mbx^i \mid \mbx_\textsc{un-masked}, \mbc) 
\end{align}
Hence, for any set $A$: 
\begin{align}
    \kl{p_\theta(\mbx^{A} \mid \mbx_\textsc{un-masked}, \mbc) \Bigg| \prod_{i\in A} p_\theta( x^i \mid \mbx_\textsc{un-masked}, \mbc)} \leq \sum_{i \in A} H(x^i \mid \mbx_\textsc{un-masked}, \mbc)
\end{align}

\paragraph{Entropy upper-bound}
Next, we prove that:
\begin{align}
    H(\mba \mid \mbr_{\textsc{un-masked}}, \mbc) \leq  H_\textsc{ub}(\mba \mid \mbr_\textsc{un-masked}, \mbc)
\end{align}
where $H_\textsc{ub}(\mba \mid \mbr_\textsc{un-masked}, \mbc) = \sum_{i} H(a^i \mid \mbr_{\textsc{un-masked}}, \mbc)$. 

Next, we note that $\kl{p_\theta(\mba \mid \mbr_{\textsc{un-masked}}, \mbc) \Bigg| \prod p_\theta(a^i \mid \mbr_{\textsc{un-masked}}, \mbc)} \geq 0$, which implies that, similar to \cref{eq:kl_entropy}, we have
\begin{align}\nonumber
   -H(\mba \mid \mbr_\textsc{un-masked}, \mbc) + H_\textsc{ub}(\mba \mid \mbr_\textsc{un-masked}, \mbc)
   &= \textsc{KL}\left(p_\theta(\mba \mid \mbr_{\textsc{un-masked}}, \mbc) \Bigg| \right. \nonumber\\
   &\qquad\left. \prod p_\theta(a^i \mid \mbr_{\textsc{un-masked}}, \mbc)\right) \\
   -H(\mba \mid \mbr_{\textsc{un-masked}}, \mbc) + H_\textsc{ub}(\mba \mid \mbr_\textsc{un-masked}, \mbc)  &\geq 0 \\
   H_\textsc{ub}(\mba \mid \mbr_\textsc{un-masked}, \mbc) &\geq H(\mba \mid \mbr_{\textsc{un-masked}}, \mbc) 
\end{align}
Note that, by \cref{eq:kl_entropy}, the gap of this bound equals the \gls{kl} divergence between the joint answer-block distribution and the product of its marginals: the bound is tight exactly when the answer tokens are conditionally independent given the unmasked reasoning and context.

\section{Fine-tuning Details}\label{appsec:finetuning}

We compare fine-tuning the LLaDA-8B Base \citep{nie2025largelanguagediffusionmodels} model on GSM8k \citep{cobbe2021training} reasoning data, versus posterior data sampled from the same model using the training questions and pre-filled answers.

\paragraph{Data}

We generate the posterior data with LLaDA-8B-Base by pre-filling the correct answer and reasoning template and sampling with entropy decoding and a block size of $128$. For the gold GSM8k training data, we preprocess the data by removing the additional computations in angle brackets, and converting the "$\#\#\#\#$" format to our reasoning template. Additionally, unlike posterior data, GSM8k reasoning traces are of varying lengths. As a result, we truncate these traces to $L=144$ tokens. When traces are longer, we truncate to the \textit{last} $144$ tokens. Approximately $\sim 12\%$ of GSM8k samples are truncated. 

\paragraph{Training}

We use a batch size of $1$ per GPU, with $8$ different noise levels per batch element.
We use LoRA \citep{hu2022lora} with $r=128, \alpha=32$. We fine-tune the model using $2$ Nvidia A100 GPUs, with a learning rate of $2.5 \times 10^{-6}$, and $32$ gradient accumulation steps. We train both models for $3300$ steps.

We modify the supervised fine-tuning code provided by \cite{zhao2025d1scalingreasoningdiffusion}.

\paragraph{Sampling}

We greedily sample from both models left-to-right with a block size of $1$. We allocate $144$ tokens for both models, and do not directly pre-fill a reasoning template.

\end{document}